*Article*# Sentiment Analysis Based on Deep Learning: A Comparative Study

**Nhan Cach Dang** [1], **María N. Moreno-García** [2] and **Fernando De la Prieta** [3,*]

1. Department of Information Technology, HoChiMinh City University of Transport (UT-HCMC), Ho Chi Minh 70000, Vietnam; tucach@hcmutrans.edu.vn
2. Data Mining (MIDA) Research Group, University of Salamanca, 37007 Salamanca, Spain; mmg@usal.es
3. Biotechnology, Intelligent Systems and Educational Technology (BISITE) Research Group, University of Salamanca, 37007 Salamanca, Spain
* Correspondence: fer@usal.es; Tel.: +34-677-522-678
Received: 31 January 2020; Accepted: 10 March 2020; Published: 14 March 2020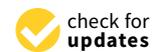

**Abstract:** The study of public opinion can provide us with valuable information. The analysis of sentiment on social networks, such as Twitter or Facebook, has become a powerful means of learning about the users' opinions and has a wide range of applications. However, the efficiency and accuracy of sentiment analysis is being hindered by the challenges encountered in natural language processing (NLP). In recent years, it has been demonstrated that deep learning models are a promising solution to the challenges of NLP. This paper reviews the latest studies that have employed deep learning to solve sentiment analysis problems, such as sentiment polarity. Models using term frequency-inverse document frequency (TF-IDF) and word embedding have been applied to a series of datasets. Finally, a comparative study has been conducted on the experimental results obtained for the different models and input features.

**Keywords:** sentiment analysis; deep learning; machine learning; neural network; natural language processing## 1. Introduction

Web 2.0 has led to the emergence of blogs, forums, and online social networks that enable users to discuss any topic and share their opinions about it. They may, for example, complain about a product that they have bought, debate current issues, or express their political views. Exploiting such information about users is key to the operation of many applications (such as recommender systems), in the survey analyses conducted by organizations, or in the planning of political campaigns. Moreover, analyzing public opinions is also very important to governments because it explains human activity and behavior and how they are influenced by the opinions of others. In the area of recommender systems and personalization, the inference of user sentiment can be very useful to make up for the lack of explicit user feedback on a provided service. In addition to machine learning, other methods, such as those based on the similarity of results, can be used for this purpose [1]. The sources of data for sentiment analysis (SA) are online social media, the users of which generate an ever-increasing amount of information. Thus, these types of data sources must be considered under the big data approach, given that additional issues must be dealt with to achieve efficient data storage, access, and processing, and to ensure the reliability of the obtained results [2].

The problem of automatic sentiment analysis (SA) is a growing research topic. Although SA is an important area and already has a wide range of applications, it clearly is not a straightforward task and has many challenges related to natural language processing (NLP). Recent studies on sentiment

*Electronics* **2020**, *9*, 483; doi:10.3390/electronics9030483      www.mdpi.com/journal/electronics

analysis continue to face theoretical and technical issues that hinder their overall accuracy in polarity detection [3,4]. Hussein et al. [4] studied the relationship between those issues and the sentiment structure, as well as their impact on the accuracy of the results. This work verifies that accuracy is a matter of high concern among the latest studies on sentiment analysis and proves that it is affected by some challenges, such as addressing negation or domain dependence.

Social media are important sources of data for SA. Social networks are continuously expanding, generating much more complex and interrelated information.

In this context, Thai et al. suggested not to focus solely on the structure and correlations of data, but on a lifelong learning approach to dealing with data presentation, analysis, inference, visualization, search and navigation, and decision making in complex networks [2].

Several studies focus on building powerful models to solve the continuously increasing complexity of big data, as well as to expand sentiment analysis to a wide range of applications, from financial forecasting [5,6] and marketing strategies [7] to medicine analysis [8,9] and other areas [10–18]. However, few of them pay attention to evaluating different deep learning techniques in order to provide practical evidence of their performance [5,17,19,20].

When examining the performance of a single method on a single dataset in a particular domain, the results show a relatively high overall accuracy [15,19,20] for Convolutional Neural Networks (CNN) and Recurrent Neural Networks (RNN). Hassan and Mahmood [15] proved that CNN and RNN models can overcome shortcoming of short text in deep learning models. Qian et al. [10] showed that Long Short-Term Memory (LSTM) behaves efficiently when used on different text levels of weather-and-mood tweets.

Li et al. [17] studied the impact of data quality on sentiment classification performance. They considered three criteria, namely informativeness, readability, and subjectivity, to assess the quality of online product reviews. The study highlighted two factors that affect the level of accuracy of sentiment analysis—readability and length of the reviews. Higher readability and shorter text datasets yielded higher quality of sentiment classification. However, when the size or domain of the data varies, the reliability of the proposed method is questionable

In comparison studies, most papers focus on reliability metrics, such as overall accuracy or F-score, and leave out processing time. In addition, the evaluations of the models are conducted on a small number of datasets. This research addresses that gap by means of a comprehensive comparison of sentiment analysis methods in the literature, and an experimental study to evaluate the performance of deep learning models and related techniques on datasets about different topics. Our research question aims to determine whether it is possible to present outperforming methods for multiple types and sizes of datasets. We build upon on previous studies of improvement of SA performance by evaluating the results from the viewpoint of a combination of three criteria: overall accuracy, F-score, and processing time. The purpose of this comparative study is to give an objective overview of different techniques that can guide researchers towards the achievement of better results

In recent years, several studies have proposed deep-learning-based sentiment analyses, which have differing features and performance. This work looks at the latest studies that have used deep learning models, such as deep neural networks (DNN), recurrent neural networks (RNN), and convolutional neural networks (CNN), to solve different problems related to sentiment analysis (e.g., sentiment polarity and aspect-based sentiment). We applied deep learning models with TF-IDF and word embedding to Twitter datasets and implemented the state-of-the-art of sentiment analysis approaches based on deep learning.

The rest of this paper is organized as follows. Section 2 provides background knowledge on this research area. Section 3 discusses related work. Section 4 describes the comparative study. Section 5 outlines the experimental results, followed by the conclusions in Section 6.

## 2. Background

*2.1. Deep Learning*

Deep learning adapts a multilayer approach to the hidden layers of the neural network. In traditional machine learning approaches, features are defined and extracted either manually or by making use of feature selection methods. However, in deep learning models, features are learned and extracted automatically, achieving better accuracy and performance. In general, the hyper parameters of classifier models are also measured automatically. Figure 1 shows the differences in sentiment polarity classification between the two approaches: traditional machine learning (Support Vector Machine (SVM), Bayesian networks, or decision trees) and deep learning. Artificial neural networks and deep learning currently provide the best solutions to many problems in the fields of image and speech recognition, as well as in natural language processing. Several types of deep learning techniques are discussed in this section.

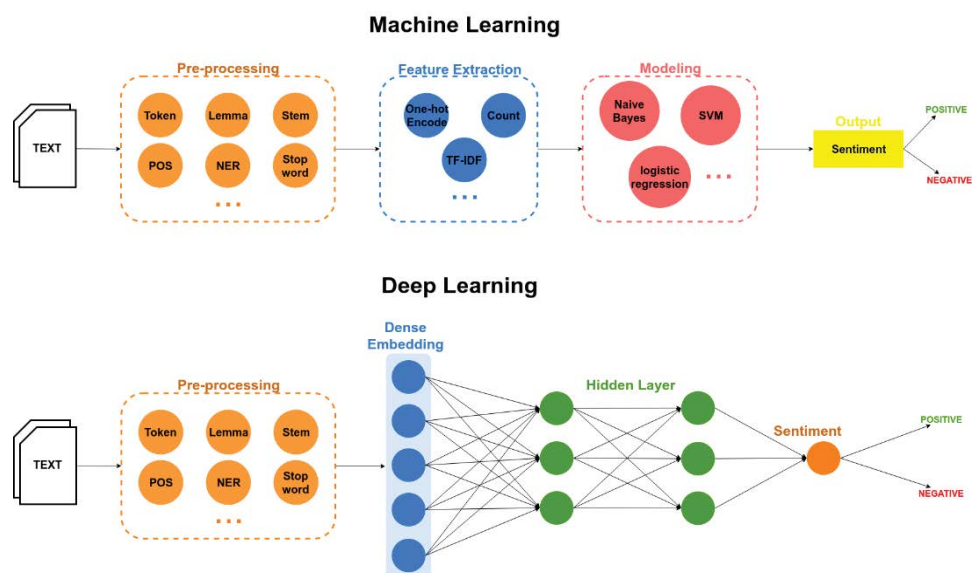

**Figure 1.** Differences between two classification approaches of sentiment polarity, machine learning (top), and deep learning (bottom). Part of Speech (POS); Named Entity Recognition (NER); Term Frequency-Inverse Document Frequency (TF-IDF).

2.1.1. Deep Neural Networks (DNN)

A deep neural network [21] is a neural network with more than two layers, some of which are hidden layers (Figure 2). Deep neural networks use sophisticated mathematical modeling to process data in many different ways. A neural network is an adjustable model of outputs as functions of inputs, which consists of several layers: an input layer, including input data; hidden layers, including processing nodes called neurons; and an output layer, including one or several neurons, whose outputs are the network outputs.

2.1.2. Convolutional Neural Networks (CNN)

A convolutional neural network is a special type of feed-forward neural network originally employed in areas such as computer vision, recommender systems, and natural language processing. It is a deep neural network architecture [22], typically composed of convolutional and pooling or subsampling layers to provide inputs to a fully-connected classification layer. Convolution layers filter their inputs to extract features; the outputs of multiple filters can be combined. Pooling or subsampling layers reduce the resolution of features, which can increase the CNN's robustness to noise and distortion. Fully connected layers perform classification tasks. An example of a CNN architecture can be seen in Figure 3. The input data was preprocessed to reshape it for the embedding matrix. The figure shows an input embedding matrix processed by four convolution layers and two max pooling layers. The first two convolution layers have 64 and 32 filters, which are used to train different features; these are followed by a max pooling layer, which is used to reduce the complexity of the output and to prevent the overfitting of the data. The third and fourth convolution layers have 16 and 8 filters, respectively, which are also followed by a max pooling layer. The final layer is a fully connected layer that will reduce the vector of height 8 to an output vector of one, given that there are two classes to be predicted (Positive, Negative).

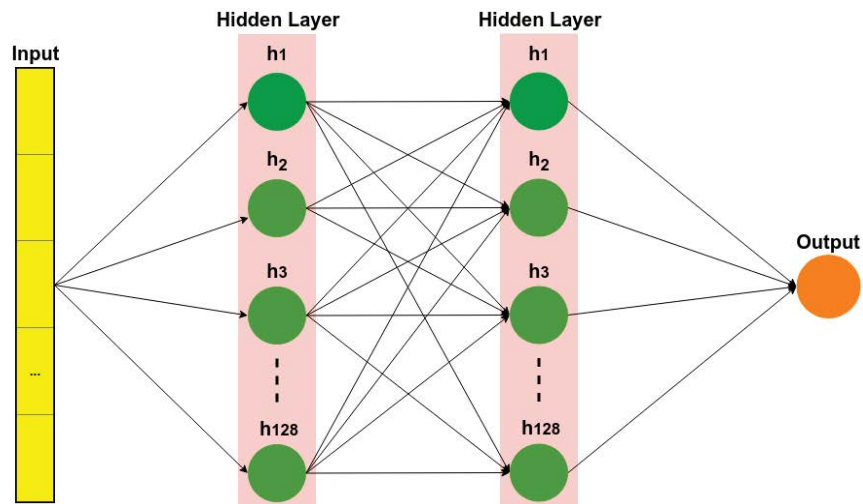

**Figure 2.** Deep neural network (DNN).

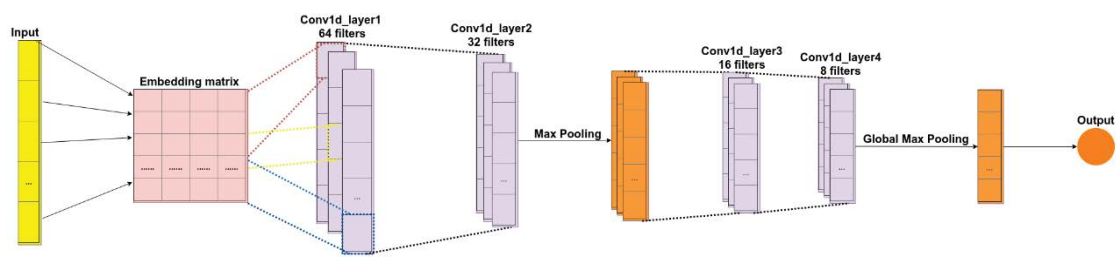

**Figure 3.** A convolutional neural network.

2.1.3. Recurrent Neural Networks (RNN)

Recurrent neural networks [23] are a class of neural networks whose connections between neurons form a directed cycle, which creates feedback loops within the RNN. The main function of RNN is the processing of sequential information on the basis of the internal memory captured by the directed cycles. Unlike traditional neural networks, RNN can remember the previous computation of information and can reuse it by applying it to the next element in the sequence of inputs. A special type of RNN is long short-term memory (LSTM), which is capable of using long memory as the input of activation functions in the hidden layer. This was introduced by Hochreiter and Schmidhuber (1997) [24]. Figure 4 illustrates an example of the LSTM architecture. The input data is preprocessed to reshape data for the embedding matrix (the process is similar to the one described for the CNN). The next layer is the LSTM, which includes 200 cells. The final layer is a fully connected layer, which includes 128 cells for text classification. The last layer uses the sigmoid activation function to reduce the vector of height 128 to an output vector of one, given that there are two classes to be predicted (positive, negative).

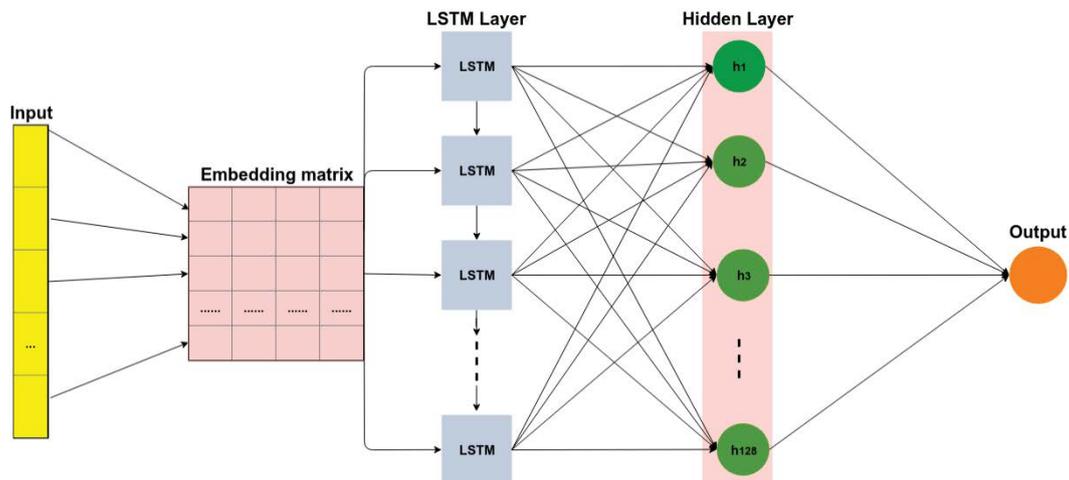

**Figure 4.** A long short-term memory network. LSTM, long short-term memory.

2.1.4. Other Neural Networks

One type of deep neural network is called a deep belief network (DBN) [25]. It comprises multiple layers of a graphical model, having both directed and undirected edges. Each network is composed of multiple layers of hidden units and each layer is connected to the next one, but the units within a layer are not connected. A DBN is learned by using a greedy layer-wise learning algorithm.

A recursive neural network (RecNN) [26] is a type of neural network that can be viewed as a generalization of RNN. Recursive neural networks are usually used to learn a directed acyclic graph structure from data. The hidden state vectors of the left and right child nodes in the graph can be used to compute for the hidden state vector of the current node.

Another category is hybrid deep learning [27], which combines two or more deep learning techniques together, such as convolutional neural networks (CNN) and long short-term memory (LSTM) [28], or probabilistic neural networks (PNN) and a two-layered restricted Boltzmann machine (RBM) [29].

*2.2. Sentiment Analysis*

Sentiment analysis is a process of extracting information about an entity and automatically identifying any of the subjectivities of that entity. The objective is to determine whether text generated by users conveys their positive, negative, or neutral opinions. Sentiment classification can be carried out on three levels of extraction: the aspect or feature level, the sentence level, and the document level. Currently, there are three approaches to address the problem of sentiment analysis [30]: (1) lexicon-based techniques, (2) machine-learning-based techniques, and (3) hybrid approaches.

**Lexicon-based** techniques were the first to be used for sentiment analysis. They are divided into two approaches: dictionary-based and corpus-based [31]. In the former type, sentiment classification is performed by using a dictionary of terms, such as those found in SentiWordNet and WordNet. Nevertheless, corpus-based sentiment analysis does not rely on a predefined dictionary but on statistical analysis of the contents of a collection of documents, using techniques based on k-nearest neighbors (k-NN) [32], conditional random field (CRF) [33], and hidden Markov models (HMM) [34], among others.

**Machine-learning-based** techniques [35] proposed for sentiment analysis problems can be divided into two groups: (1) traditional models and (2) deep learning models. Traditional models refer to classical machine learning techniques, such as the naïve Bayes classifier [36], maximum entropy classifier [37,38], or support vector machines (SVM) [39]. The input to those algorithms includes lexical features, sentiment lexicon-based features, parts of speech, or adjectives and adverbs. The accuracy of these systems depends on which features are chosen. Deep learning models can provide better results than traditional models. Different kinds of deep learning models can be used for sentiment analysis, including CNN, DNN, and RNN. Such approaches address classification problems at the document level, sentence level, or aspect level. These deep learning methods will be discussed in the following section.

The **hybrid approaches** [40] combine lexicon- and machine-learning-based approaches. Sentiment lexicons commonly play a key role within a majority of these strategies. Figure 5 illustrates a taxonomy of deep-learning-based methods for sentiment analysis.

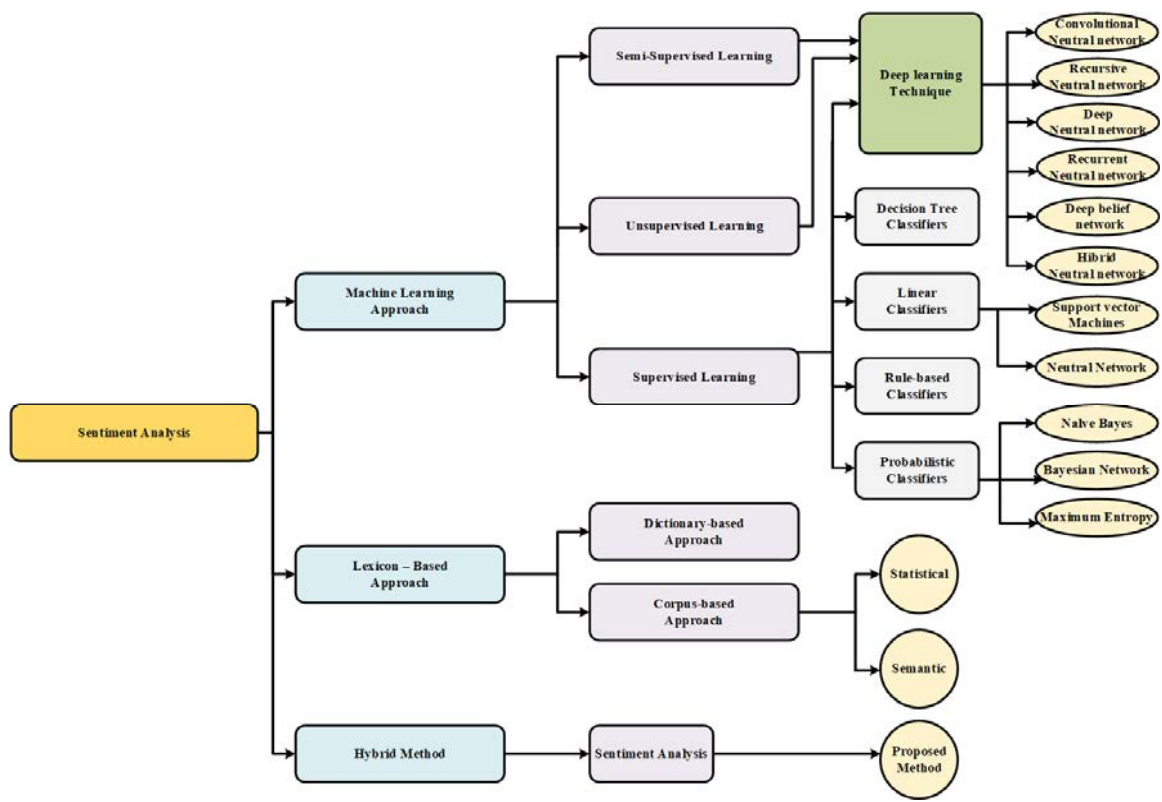

**Figure 5.** Taxonomy of sentiment analysis techniques. Source: [30,41].

Sentiment analysis, whether performed by means of deep learning or traditional machine learning, requires that text training data be cleaned before being used to induce the classification model. Tweets usually contain white spaces, punctuation marks, non-characters, Retweet (RT), "@ links", and stop words. These characters could be removed using libraries such as BeautifulSoup because they do not contain any information that would be useful for sentiment analysis. After cleaning, tweets can be split into individual words, which are transformed into their base form by lemmatization, then converted into numerical vectors by using methods such as word embedding or term frequency-inverse document frequency (TF-IDF).

Word embedding [42] is a technique for language modeling and feature learning, where each word is mapped to a vector of real values in such a way that words with similar meanings have a similar representation. Value learning can be done using neural networks. A commonly used word embedding system is Word2vec (GloVe, or Gensim), which contains models such as skip-gram and continuous bag-of-words (CBOW). Both models are based on the probability of words occurring in proximity to each other. Skip-gram makes it possible to start with a word and predict the words that are likely to surround it. Continuous bag-of-words reverses that by predicting a word that is likely to occur on the basis of specific context words.

TF-IDF is a statistical measure reflecting how important a word is to a document in a collection or corpus. This metric considers the frequency of the word in the target document, as well as the frequency in the other documents of the corpus. The higher the frequency of a word in a target document and the lower its frequency in other documents, the greater its importance. The vectorizer class in the scikit-learn library is usually used to compute TF-IDF.

Both word embedding and TF-IDF are used as input features of deep learning algorithms in NLP. Sentiment analysis tasks transform collections of raw data into vectors of continuous real numbers.

There are different kinds of tasks, such as objective or subjective classification, polarity sentiment detection, and feature- or aspect-based sentiment analysis. The subjectivity of words and phrases may depend on their context and an objective document may contain subjective sentences. Aspect-based sentiment analysis refers to sentiments expressed towards specific aspects of entities (e.g., value, room, location, cleanliness, or service). Polarity and intensity are two components used to score sentiment analysis. Polarity indicates whether the sentiment is negative, neutral, or positive. Intensity indicates the relative strength of the sentiment.

*2.3. Application of Sentiment Analysis*

It is widely accepted that sentiment analysis is very useful in a wide range of application domains, such as business, government, and biomedicine.

In the fields of business intelligence and e-commerce, companies can study customers' feedback to provide better customer support, build better products, or improve their marketing strategies to attract new customers. Sentiment analysis can be used to infer the users' opinions on events or products. The results of SA help to gain greater insight into the customers' interests or opinions on industrial trends. In this context, Jain and Dandannavar [43] proposed a fast, flexible, and scalable SA framework for sentiment analysis of Twitter data that involves the use of some machine learning methods and Apache spark.

As pointed out in the introduction, the area of recommender systems has also benefited from sentiment analysis. A sample of this can be found in the work of Preethi et al. [12], where recursive neural networks were applied to analyze sentiments in reviews. The output was used to improve and validate the restaurant and movie recommendations of a cloud-based recommender system. Along with behavioral analysis, sentiment analysis is also an efficient tool for commodity markets [7].

The medical domain is another field of potential interest. The applications of opinion mining in health-related texts on social media and blogs were explored in [8]. In addition to traditional machine learning and text processing techniques, the author offers new approaches and proposes a medical lexicon to support experts and patients in the varied methodology that is used to describe symptoms and diseases. In the field of mental health, sentiment analysis is performed on texts written by patients' posts on social media as a means of supplementing or replacing the questionnaires they usually fill in [9].

**3. Related Work**

The purpose of this study is to review different approaches and methods in sentiment analysis that can be taken as a reference in future empirical studies. We have focused on key aspects of research, such as technical challenges, datasets, the methods proposed in each study, and their application domains.

Recently, deep learning models (including DNN, CNN, and RNN) have been used to increase the efficiency of sentiment analysis tasks. In this section, state-of-the-art sentiment analysis approaches based on deep learning are reviewed.

Beginning in 2015, many authors have since evaluated this trend. Tang et al. [44] introduced techniques based on deep learning approaches for several sentiment analyses, such as learning word embedding, sentiment classification, and opinion extraction. Zhang and Zheng [35] discussed machine learning for sentiment analysis. Both research groups used part of speech (POS) as a text feature and used TF-IDF to calculate the weight of words for the analysis. Sharef et al. [45] discussed the opportunities of sentiment analysis approaches for big data. In papers [13,18,46], the latest deep-learning-based techniques (namely CNN, RNN, and LSTM) were reviewed and compared with each other in the context of sentiment analysis problems.

Some other studies applied deep-learning-based sentiment analysis in different domains, including finance [5,6], weather-related tweets [10], trip advisors [11], recommender systems for cloud services [12], and movie reviews [13–18]. In [10], where text features were automatically extracted from different data sources, user information and weather knowledge were transferred into word embedding using the Word2vec tool. The same techniques have been used in several works [5,47]. Jeong et al. [48] identified product development opportunities by combining topic modeling and the results of a sentiment analysis that had been performed on customer-generated social media data. It has been used as a real-time monitoring tool for analysis of changing customer needs in rapidly evolving product environments. Pham et al. used multiple layers of knowledge representation to analyze travel reviews and determine sentiments for five aspects, including value, room, location, cleanliness, and service [11]. Another approach [49] combines sentiment and semantic features in an LSTM model based on emotion detection. Preethi et al. [12] applied deep learning to sentiment analysis for a recommender system in the cloud using the food dataset from Amazon. For the health domain, Salas-Zárate et al. [31] applied an ontology-based, aspect-level sentiment analysis method to tweets about diabetes.

Polarity-based sentiment deep learning applied to tweets was found in [19,20,28,36,40,50]. The authors described how they used deep learning models to increase the accuracy of their respective sentiment analysis. Most of the models are used for content written in English, but there are a few that manage tweets in other languages, including Spanish [51], Thai [28], and Persian [47]. Previous researchers have analyzed tweets by applying different models of polarity-based sentiment deep learning. Those models include DNN [50], CNN [20], and hybrid approaches [40].

Other works using neural network models are focused not only on the sentiment polarity of textual content, but also on aspect sentiment analysis [6,11,31,52–54]. Salas-Zárate et al. [31] used semantic annotation (diabetes ontology) to identify aspects from which they performed aspect-based sentiment analysis using SentiWordNet. Pham et al. [11] included the determination of sentiment ratings and importance degrees of product aspects. A novel, multilayer architecture was proposed to represent customer reviews aiming at extracting more effective sentiment features.

From among 32 of the analyzed studies, we identified three popular models for sentiment polarity analysis using deep learning: DNN [50], CNN [20], and hybrid [40]. In [13,18,46], three deep learning techniques, namely CNN, RNN, and LSTM, were individually tested on different datasets. However, there was a lack of a comparative analysis of these three techniques.

Many studies use the same process for sentiment analysis. First, text features are automatically extracted from different data sources, then they are transferred into word embedding using the Word2vec tool [5,10,47].

Sentiment analysis has also been the target of extensive research in the application domain of recommender systems. Most methods in this area are based on information filtering, and they can be classified into four categories: content-based, collaborative filtering (CF), demographic-based, and hybrid. Social media data can be used with these techniques in different ways. Content-based methods make use of characteristics of items and user's profiles, CF methods require implicit or explicit user preferences, demographic methods exploit user demographic information (age, gender, nationality, etc.), and hybrid approaches take advantage of any kind of item and user information that can be extracted or inferred from social media (actions, preferences, behavior, etc.).

Besides, when dealing with both explicit data (which are provided directly by users) and implicit data (which are inferred from the behavior and actions of users), hybrid methods and lifelong learning algorithms are considered as in-depth approaches for recommendation systems.

Shoham [55] proposed one of the first hybrid recommendation systems, which takes advantage of both content and collaborative filtering recommendation methods. The content-based part of the proposal involves the identification of user profiles based on their interest in topics extracted from web pages, while the collaborative filtering part of the system is based on the feedback of other users.

Although sentiment analysis is not performed in this work, it can be considered the precursor of other studies combining both approaches in which sentiment analysis is used to obtain implicit user feedback. A recent study from Wang et al. [56] presents a hybrid approach in which sentiment analysis of reviews about movies is used in order to improve a preliminary recommendation list obtained from the combination of collaborative filtering and content-based methods. In the same application domain, Singh et al. propose the use of a sentiment classifier induced from movie reviews as a second filter after collaborative filtering [57].

In addition, one advanced machine learning paradigm is the so-called holistic models or lifelong learning algorithms, which are argued to significantly improve sentiment analysis accuracy [58]. While other methods learn a model by using only data for a particular application, this method attains a continually updating knowledge base of attributes, such as sentiment polarity or sentiment aspects. Stai et al. [59] introduced a social recommendation framework, whose main objective is the creation of enriched multimedia content adapted to users. This is achieved through a holistic approach, where the explicit and implicit relevance feedback from users is derived from their interactions with both the video and its enrichment. Although this method represents a significant improvement over other approaches, it requires personal user information.

Table 1 summarizes 32 important papers related to our research. It includes the year of publication, authors' names, research work, methods, datasets, and the study target.

## 4. Comparative Study

In this section, we begin by introducing different topics pertaining to datasets, and then we offer details about the sentiment classification process.

We used eight datasets in our experiments on sentiment polarity analysis. Three of them contain tweets; the largest has 1.6 million tweets, with each one labeled as either positive or negative sentiment, while the other two datasets contain 14,640 and 17,750 tweets, respectively, labeled as positive, negative, or neutral. The remaining five datasets include a total of 125,000 comments from user reviews of movies, books, and music labeled as either positive or negative sentiments.

Two approaches for preparing inputs to the classification algorithms are compared in our experiments: word embedding and TF-IDF. For word embedding, we applied Word2vec, which contains models such as skip-gram and continuous bag-of-words (CBOW). Skip-gram makes it possible to start with a word and predict the words that are likely to surround it. Continuous bag-of-words reverses that and enables the prediction of a word that is likely to occur in the context of words. For TF-IDF, we used the vectorizer class in the scikit-learn library.

We conducted an experimental study where three models (DNN, CNN, and RNN) were trained and evaluated on different datasets, which had been preprocessed with both word embedding and TF-IDF. The objective was to compare the performance of all these techniques and improve the state-of-the-art of sentiment analysis tasks.

Table 1. Summary of deep-learning-based sentiment analysis.

| No. | Year | Study | Research Work | Method | Dataset | Target |
|---|---|---|---|---|---|---|
| 1 | 2019 | Alharbi el al. [19] | Twitter sentiment analysis | CNN | SemEval 2016 workshop | Feature extraction from user behavior information |
| 2 | 2019 | Kraus et al. [16] | Sentiment analysis based on rhetorical structure theory | Tree-LSTM and Discourse-LSTM | Movie Database (IMD), food reviews (Amazon) | Aim to improve accuracy |
| 3 | 2019 | Do et al. [53] | Comparative review of sentiment analysis based on deep learning | CNN, LSTM, GRU, and hybrid approaches | SemEval workshop and social network sites | Aspect extraction and sentiment classification |
| 4 | 2019 | Abid et al. [20] | Sentiment analysis through recent recurrent variants | CNN, RNN | Twitter | Domain-specific word embedding |
| 5 | 2019 | Yang et al. [52] | Aspect-based sentiment analysis | Coattention-LSTM, Coattention-MemNet, Coattention-LSTM + location | Twitter, SemEval 2014 | Target-level and context-level feature extraction |
| 6 | 2019 | Wu et al. [60] | Sentiment analysis with variational autoencoder | LSTM, Bi-LSTM | Facebook, Chinese VA, Emobank | Encoding, sentiment prediction, and decoding |
| 7 | 2018 | Pham et al. [11] | Aspect-based sentiment analysis | LRNN-ASR, FULL-LRNN-ASR | Tripadvisor | Enriching knowledge of the input through layers |
| 8 | 2018 | Sohangir et al. [5] | Deep learning for financial sentiment analysis | LSTM, doc2vec, and CNN | StockTwits | Improving the performance of sentiment analysis for StockTwits |
| 9 | 2018 | Li et al. [17] | How textual quality of online reviews affect classification performance | SRN, LSTM, and CNN | Movie reviews from imdb.com | Impact of two influential textual features, namely the word count and review readability |
| 10 | 2018 | Zhang et al. [61] | Textual sentiment analysis via three different attention convolutional neural networks and cross-modality consistent regression | CNN | SemEval 2016, Sentiment Tree Bank | LSTM attention and attentive pooling is integrated with CNN model to extract sentence features based on sentiment embedding, lexicon embedding, and semantic embedding |

Table 1. *Cont.*

| No. | Year | Study | Research Work | Method | Dataset | Target |
|---|---|---|---|---|---|---|
| 11 | 2018 | Schmitt et al. [54] | Joint aspect and polarity classification for aspect-based sentiment analysis | CNN, LSTM | SemEval 2017 | Approach based on aspect sentiment analysis to solve two classification problems (aspect categories + aspect polarity) |
| 12 | 2018 | Qian et al. [10] | Sentiment analysis model on weather-related tweets | DNN, CNN | Twitter, social network sites | Feature extraction |
| 13 | 2018 | Tang et al. [62] | Improving the state-of-the-art in many deep learning sentiment analysis tasks | CNN, DNN, RNN | Social network sites | Sentiment classification, opinion extraction, fine-grained sentiment analysis |
| 14 | 2018 | Zhang et al. [22] | Survey of deep learning for sentiment analysis | CNN, DNN, RNN, LSTM | Social network sites | Sentiment analysis with word embedding, sarcasm analysis, emotion analysis, multimodal data for sentiment analysis |
| 15 | 2017 | Choudhary et al. [30] | Comparative study of deep-learning-based sentimental analysis with existing techniques | CNN, DNN, RNN, lexicon, hybrid | Social network sites | Domain dependency, sentiment polarity, negation, feature extraction, spam and fake review, huge lexicon, bi-polar words |
| 16 | 2018 | Jangid et al. [6] | Financial sentiment analysis | CNN, LSTM, RNN | Financial tweets | Aspect-based sentiment analysis |
| 17 | 2017 | Araque et al. [63] | Enhancing deep learning sentiment analysis with ensemble techniques in social applications | Deep-learning-based sentiment classifier using a word embedding model and a linear machine learning algorithm | SemEval 2013/2014, Vader, STS-Gold, IMDB, PL04, and Sentiment140 | Improving the performance of deep learning techniques and integrating them with traditional surface approaches based on manually extracted features |
| 18 | 2017 | Jeong et al. [48] | A product opportunity mining approach based on topic modeling and sentiment analysis | LDA-based topic modeling, sentiment analysis, and opportunity algorithm | Twitter, Facebook, Instagram, and Reddit | Identification of product development opportunities from customer-generated social media data |
| 19 | 2017 | Gupta et al. [49] | Sentiment-/semantic-based approaches for emotion detection | LSTM-based deep learning | Twitter | Combining sentiment and semantic features |

Table 1. *Cont.*

| No. | Year | Study | Research Work | Method | Dataset | Target |
|---|---|---|---|---|---|---|
| 20 | 2017 | Preethi et al. [12] | Sentiment analysis for recommender system in the cloud | RNN, naïve Bayes classifier | Amazon | Recommending the places that are near to the user's current location by analyzing the different reviews and consequently computing the score grounded on it |
| 21 | 2017 | Ramadhani et al. [50] | Twitter sentiment analysis | DNN | Twitter | Handling a huge amount of unstructured data |
| 22 | 2017 | Ain et al. [13] | A review of sentiment analysis using deep learning techniques | CNN, RNN, DNN, DBN | Social network sites | Analyzing and structuring hidden information extracted from social media in the form of unstructured data |
| 23 | 2017 | Roshanfekr et al. [47] | Sentiment analysis using deep learning on Persian texts | NBSVM-Bi, Bidirectional-LSTM, CNN | Customer reviews from www.digikala.com | Evaluating deep learning methods using the Persian language |
| 24 | 2017 | Paredes-Valverde et al. [51] | Sentiment analysis for improvement of products and services | CNN + Word2vec | Twitter in Spanish | Detecting customer satisfaction and identifying opportunities for improvement of products and services |
| 25 | 2017 | Jingzhou Liu et al. [64] | Extreme multilabel text classification | XML-CNN | RCV1, EUR-Lex, Amazon, and Wiki | Capturing richer information from different regions of the document |
| 26 | 2017 | Hassan et al. [15] | Sentiment analysis of short texts | CNN, LSTM, on top of pretrained word vectors | Stanford Large Movie Review, IMDB, Stanford Sentiment Treebank, SSTb | Achieving comparable performances with fewer parameters on sentiment analysis tasks |

Table 1. *Cont.*

| No. | Year | Study | Research Work | Method | Dataset | Target |
|---|---|---|---|---|---|---|
| 27 | 2017 | Chen et al. [65] | Multimodal sentiment analysis with word-level fusion and reinforcement learning | Gated multimodal embedding LSTM with temporal attention | CMU-MOSI | Developing a novel deep architecture for multimodal sentiment analysis that performs modality fusion at the word level |
| 28 | 2017 | Al-Sallab et al. [66] | Opinion mining in Arabic as a low-resource language | Recursive deep learning | Online comments from QALB, Twitter, and Newswire articles written in MSA | Providing more complete and comprehensive input features for the autoencoder and performing semantic composition |
| 29 | 2016 | Vateekul et al. [28] | A study of sentiment analysis in Thai | LSTM, DCNN | Twitter | Finding the best parameters of LSTM and DCNN |
| 30 | 2016 | Singhal, et al. [18] | A survey of sentiment analysis and deep learning | CNN, RNTN, RNN, LSTM | Sentiment Treebank dataset, movie reviews, MPQA, and customer reviews | Comparison of classification performance of different models on different datasets |
| 31 | 2016 | Gao et al. [14] | Sentiment analysis using AdaBoost combination | CNN | Movie reviews and IMDB | Studying the possibility of leveraging the contribution of different filter lengths and grasping their potential in the final polarity of the sentence |
| 32 | 2016 | Rojas-Barahona et al. [46] | Overview of deep learning for sentiment analysis | CNN, LSTM | Movie reviews, Sentiment Treebank, and Twitter | To extract the polarity from the data |

Gated Recurrent Units (GRU); Bi-directional Long-Short-Term-Memory (Bi-LSTM); Latent Rating Neural Network-Aspect Semantic Representation (LRNN-ASR); Simple Recurrent Networks (SRN); Latent Dirichlet Allocation (LDA); Naive Bayes and Support Vector Machine Bidirectional (NBSVM-bi); Deep Convolutional Neural Network (DCNN); Recursive Neural Tensor Network (RNTN); Multi-Perspective Question Answering (MPQA); Multimodal Opinion Sentiment Intensity (CMU-MOSI); Qatar Arabic Language Bank (QALB)

*4.1. Datasets*

Studies that perform sentiment analyses either generate their own data or use available datasets. Generating a new dataset makes it possible to use data that fits the problem the analysis is targeted at; moreover, the use of personal data ensures that no privacy laws are violated [67]. However, the main drawback is having to label the dataset, which is a challenging task. Moreover, it is not always easy to generate a large volume of data. Our approach to selecting datasets was based on their availability and accessibility. Respecting personal privacy was another factor that was considered, given that it appears in the regulations of most journals as a requirement for article publication.

Thus, we carefully chose datasets that are widely accepted by the research community.

In addition, one of our main concerns was the extensibility of the results obtained in the study. Therefore, the datasets were obtained from different sources and they cover different topics in order to perform a wide range of experiments. In this way, the results have made it possible to make a comprehensive comparison of the performance of deep learning models in sentiment analysis. We also considered the size of the datasets; the larger they are, the more possibilities they offer, even though this also increases their complexity. We worked with labeled datasets from which personal information was removed, since this information was not needed to test the performance of sentiment analysis models. These datasets are described below:

- Sentiment140 was obtained from Stanford University [68]. It contains 1.6 million tweets about products or brands. The tweets were already labeled with the polarity of the sentiment conveyed by the person writing them (0 = negative, 4 = positive).
- Tweets Airline [69] is a tweet dataset containing user opinions about U.S. airlines. It was crawled in February 2015. It has 14,640 samples, and it was divided into negative, neutral, and positive classes.
- Tweets SemEval [70] is a tweet dataset that includes a range of named geopolitical entities. This dataset has 17,750 samples, and it was divided into positive, neutral, and negative classes.
- IMDB Movie Reviews [71] is a dataset of comments from audiences about the stories in films. It has 25,000 samples divided into positive and negative.
- IMDB Movie Reviews was obtained from Stanford University [72]. This dataset contains comments from audiences about the story of films. It has 50,000 samples, which are divided into positive and negative.
- Cornell Movie Reviews [73] contains comments from audiences about the stories in films. This dataset includes 10,662 samples for training and testing, which are labeled negative or positive.
- Book Reviews and Music Reviews is a dataset obtained from the Multidomain Sentiment of the Department of Computer Science of Johns Hopkins University. Biographies, Bollywood, Boom Boxes, and Blenders: Domain Adaptation for Sentiment Classification [74] contains user comments about books and music. Each has 2,000 samples with two classes—negative and positive.

Figure 6 shows an original sample of tweets in one of the datasets. It contains information on each of the following fields:

- "target" is the polarity of the tweet;
- "id" is the unique ID of each tweet;
- "date" is the date of the tweet;
- "query_string" indicates whether the tweet has been collected with any particular query keyword (for this column, 100% of the entries labeled are with the value "NO_QUERY");
- "user" is the Twitter handle name of the user who tweeted;
- "text" is the verbatim text of the tweet.

We used the "text" and "target" fields to perform the experiment.

Electronics **2020**, *9*, 483    15 of 29**Figure 6.** Examples of the Sentiment140 dataset.

*4.2. Methodological Approach*

After reviewing the proposed sentiment analysis methods in Section 3, we identified three popular approaches that have been used frequently in recent studies, namely DNN, CNN, and RNN. These models have been employed in the majority of the 32 reviewed papers, have been widely tested, and provide highly accurate results when working with different types of datasets [75]. However, no comparative study involving those algorithms has been conducted.

The focus of this research was the deep learning approach; therefore, we performed a comparative study of the performance of the three most popular deep learning models (DNN, CNN, and RNN) on eight datasets. Moreover, two text processing techniques (word embedding and TF-IDF) were employed in data preprocessing. The objective of the experiments is to compare the performance of these techniques, contributing in this way to the state-of-the-art literature on sentiment analysis tasks. These algorithms were applied to predict the sentiment polarity of the text and classify it according to that polarity. The performance of those methods was evaluated by means of the most suitable metrics used for classification problems: overall accuracy, recall, F-score, and Area Under Curve (AUC). We used k-fold cross validation with k = 10 in the application of the metrics. More details about the application of DNN, CNN, and RNN algorithms with word embedding and TF-IDF are given in Section 2.

*4.3. Sentiment Classification*

The process of sentiment analysis is discussed below. Data cleaning and feature extraction were performed in the preprocessing stages. In the training stage, several deep learning models were used. Detailed results are presented in the next section.

The main objective of our study is to evaluate the deep learning models. We used k-fold cross validation with k = 10 to determine the performance of the algorithms. All of them were tested with word embedding and TF-IDF.

Text cleaning is a preprocessing step that removes words or other components that do not contain relevant information, and thus may reduce the effectiveness of sentiment analysis. Text or sentence data include white space, punctuation, and stop words. Text cleaning has several steps for sentence normalization. All datasets were cleaned using the following steps:

- Cleaning the Twitter RTs, @, #, and the links from the sentences;
- Stemming or lemmatization;
- Converting the text to lower case;
- Cleaning all the non-letter characters, including numbers;
- Removing English stop words and punctuation;
- Eliminating extra white spaces;
- Decoding HTML to general text.

A certain processing method was then performed depending on the dataset to facilitate model formation. For example, for the Sentiment140 dataset, we dropped the columns that are not useful for sentiment analysis purposes: {"id", "date", "query_string", "user"} and converted class label values {4, 0} to {1, 0} (1 = positive, 0 = negative). For the Tweets Airline and Tweets SemEval datasets, we removed all samples labeled "neutral", leaving only two classes for the experiment—positive and negative.

After the datasets were cleaned, sentences were split into individual words, which were returned to their base form by lemmatization. At this point, sentences were converted into vectors of continuous real numbers (also known as feature vectors) by using two methods: word embedding and TF-IDF. Both kinds of feature vectors were the inputs for the deep learning algorithms evaluated in the study. Those algorithms were CNN, DNN, and RNN. Thus, two models were induced per algorithm, one for each type of vector.

*4.4. Sentiment Model*

Most traditional models use well-known features, such as bag-of words, n-grams, and TF-IDF. Such features do not consider the semantic similarity between words. Currently, many deep learning models in NLP require word embedding results as input features. Figure 7 shows the semantic similarity of the words that are the closest to "iPhone", "Obama", and "university". The words nearest to "Obama" are "president", "leader", and "election". The words nearest to "university" are "students", "education", and "master". Since neural networks can be deployed to solve sentiment classification using word embedding, we use Word2vec to train initial word vectors from the datasets that were described above.

Figure 8 shows word clouds produced from some of the topics of the datasets described in Section 4.1. These datasets were cleaned before being transformed into vectors. The figure demonstrates how topics can be easily identified. The book topic is shown in the top left corner, the movie topic is shown in the top right corner, the left bottom corner shows the music topic, and finally the airplane topic is shown in the bottom right corner.

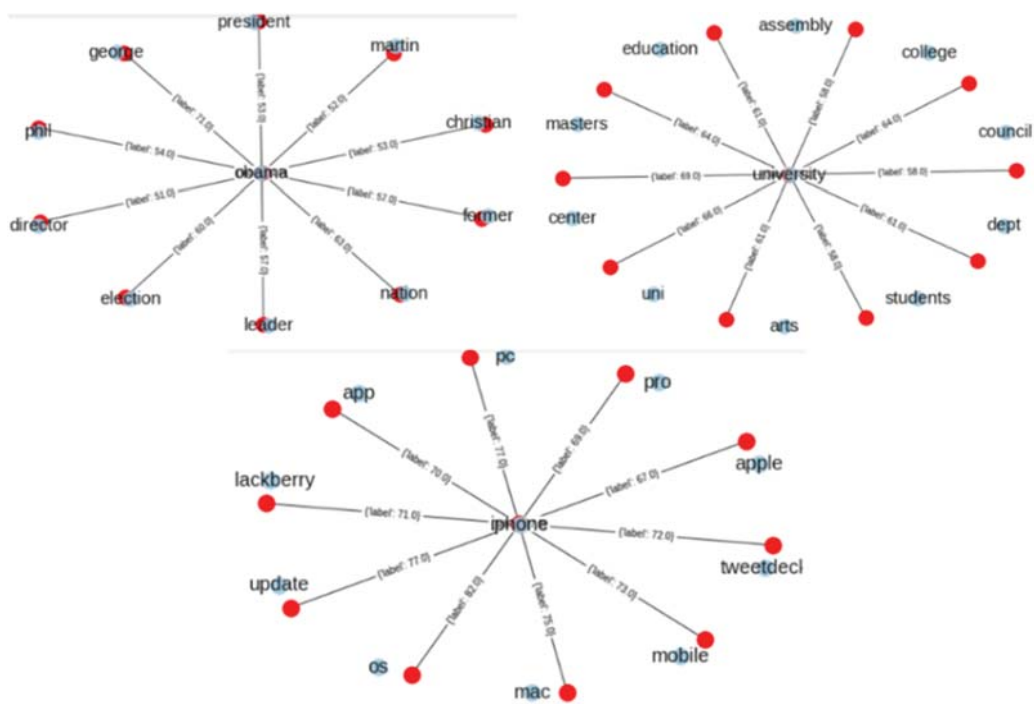

**Figure 7.** Word embedding after training the dataset with a minimum count of 5000.

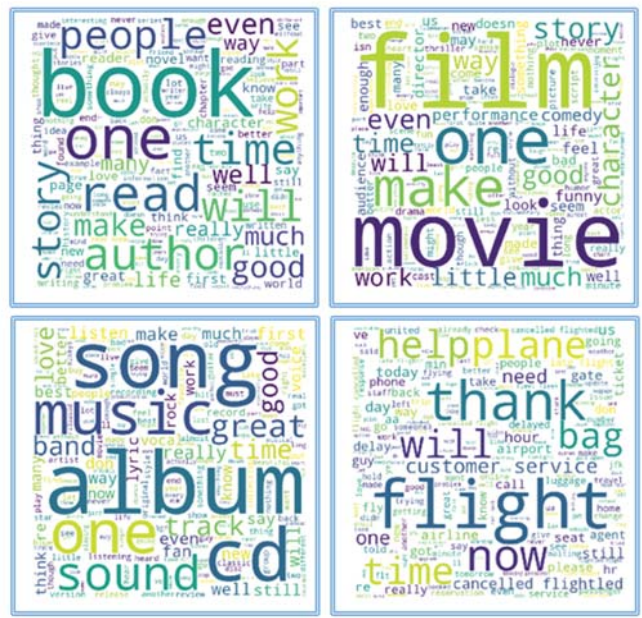

**Figure 8.** Word cloud view of the topics of the datasets.

As stated before, we used k-fold cross validation to determine the effectiveness of different embedding with k = 10. The details are shown in the experimental results section. Figure 9 shows the details of the CNN model, which are explained below.

```
Layer (type)                 Output Shape              Param #
=================================================================
embedding_1 (Embedding)      (None, 40, 300)           4500300
_________________________________________________________________
conv1d_1 (Conv1D)            (None, 40, 64)            57664
_________________________________________________________________
conv1d_2 (Conv1D)            (None, 40, 32)            6176
_________________________________________________________________
max_pooling1d_1 (MaxPooling1 (None, 13, 32)            0
_________________________________________________________________
conv1d_3 (Conv1D)            (None, 13, 16)            1552
_________________________________________________________________
conv1d_4 (Conv1D)            (None, 13, 8)             264
_________________________________________________________________
global_average_pooling1d_1 ( (None, 8)                 0
_________________________________________________________________
dense_1 (Dense)              (None, 1)                 9
=================================================================
Total params: 4,565,965
Trainable params: 65,665
Non-trainable params: 4,500,300
_________________________________________________________________
```

**Figure 9.** The details of the CNN model, which was set up for the experiment.

The function embedding is the embedding layer that is initialized with random weights and which will learn the embedding for all words in the training datasets. In our case, the size of the vocabulary is 15,000, the output dim is 300, and the maximum length is 40. The results are in a 40 × 300 matrix.

The first 1D CNN layer defines a filter of kernel size 3. For this, we will define 64 filters. This allows us to train 64 different features on the first layer of the network. Thus, the output of the first neural network layer is a 40 × 64 neuron matrix, and the result from the first CNN will be fed into the second

CNN layer. We will again define 32 different filters to be trained on this level. Following the same logic as the first layer, the output matrix will measure 40 × 32.

The maximum pooling layer is often used after a CNN layer in order to reduce the complexity of the output and prevent overfitting of the data. In our case, we choose a size of three. This means that the size of the output matrix of this layer is 13 × 32.

The third and fourth 1D CNN layers are in charge of learning higher level features. The outputs of those two layers are a 13 × 16 matrix and a 13 × 8 matrix.

The average pooling layer is a pooling layer used to further avoid overfitting. We will use the average value instead of the maximum value because it will give better results in this case. The output matrix has a size of 1 × 8 neurons.

The fully connected layer with sigmoid activation is the final layer that will reduce the vector of height 8 to 1 for prediction ("positive", "negative").

## 5. Experimental Results

To conduct the tests, we used a GeForce GTX2070 GPU card, and the Keras (https://keras.io) and Tensorflow (https://www.tensorflow.org/) libraries. DNN, CNN, and RNN models were applied to perform experiments with the different datasets described above, in order to analyze the performance of those algorithms using both word embedding and TF-IDF feature extraction.

In all the experiments, we configure the parameter for our code, such as echoes = 5, batch size = 4096, and k-fold = 10.

Accuracy, AUC, and F-score were the metrics used to evaluate the performance of the models through all experiments. Since F-score is derived from recall and precision, we also show these two measures for reference purposes.

Sentiment140 was the first dataset to be processed. Its contents were labeled as positive or negative. Since this dataset contains a much larger number of tweets than the other datasets, we first analyzed the performance of the models induced from different subsets formed with different percentages of the initial data, ranging from 10% to 100%. As shown in Figures 10–15, the combinations of feature extractors and deep learning techniques applied to those subsets produced different results for Sentiment140 data.

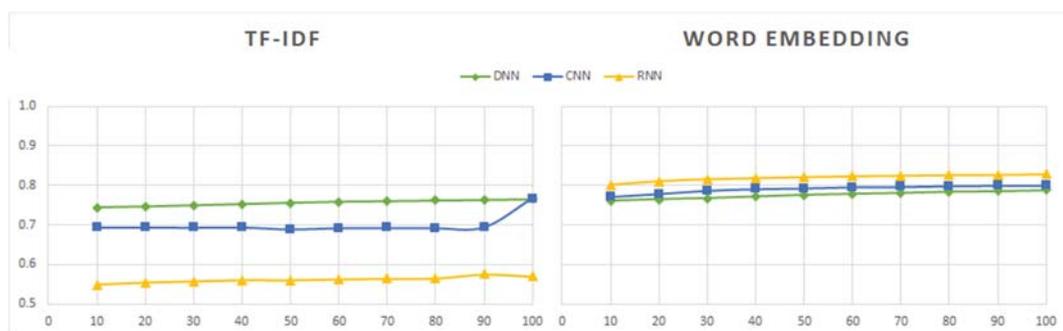

**Figure 10.** Accuracy values of deep-learning models with TF-IDF and word embedding for different numbers of tweets (percentage of the dataset).

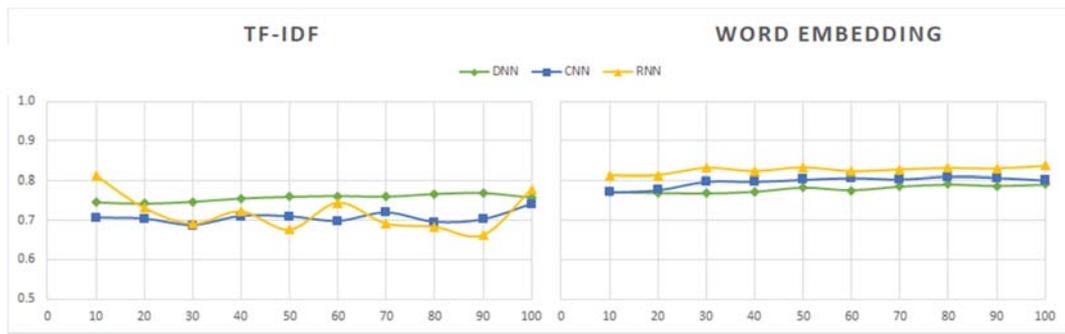

**Figure 11.** Recall values of DNN, CNN, and RNN models with TF-IDF and word embedding for different numbers of tweets (percentage of the dataset).

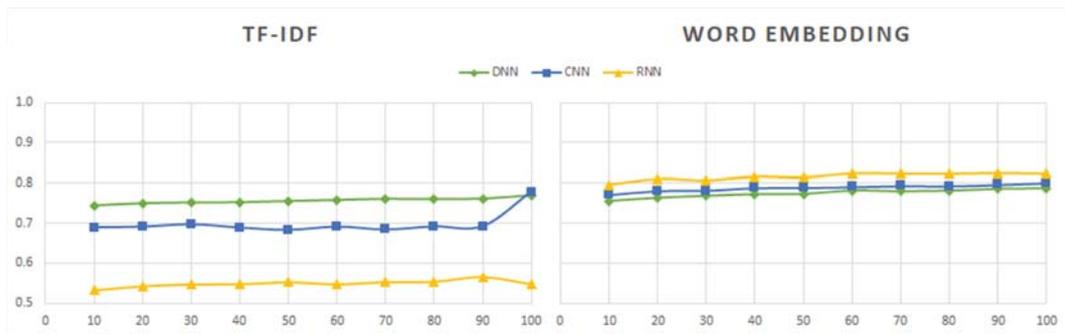

**Figure 12.** Precision values of DNN, CNN, and RNN models with TF-IDF and word embedding for different numbers of tweets (percentage of the dataset).

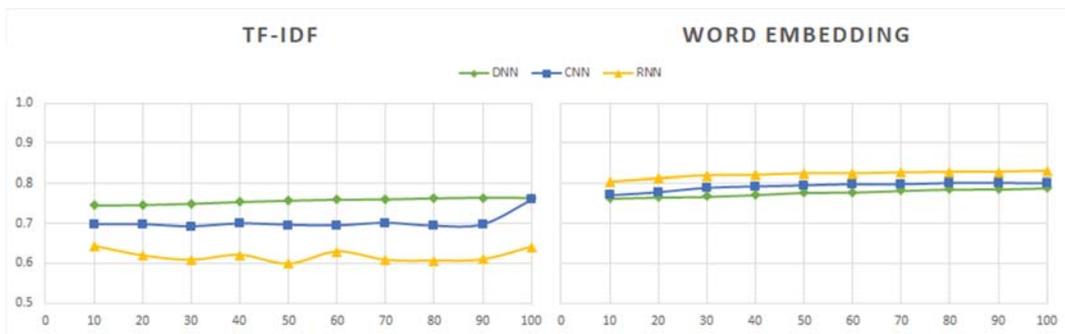

**Figure 13.** F-score values of DNN, CNN, and RNN models with TF-IDF and word embedding for different numbers of tweets (percentage of the dataset).

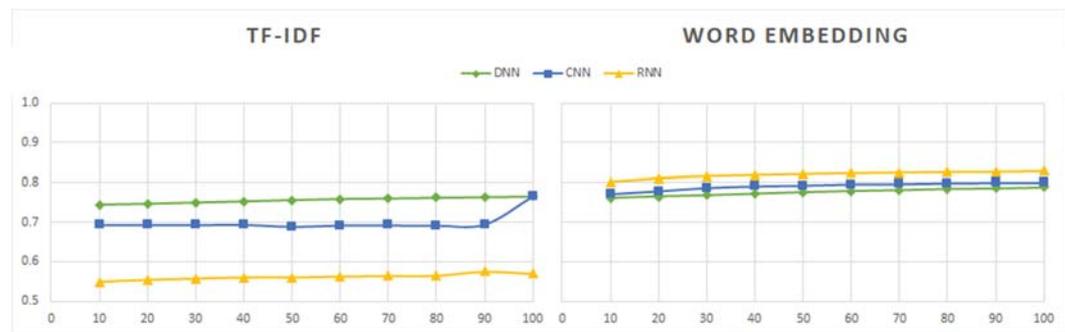

**Figure 14.** AUC values of DNN, CNN, and RNN models with TF-IDF and word embedding for different numbers of tweets (percentage of the dataset).

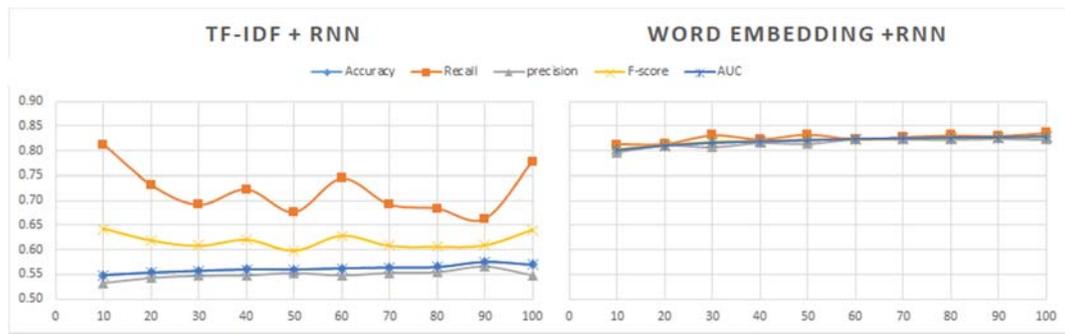

**Figure 15.** Comparison of all measures for RNN with TF-IDF (left) and word embedding (right).

An observation that is clearly seen in Figures 10–15 is the best behavior of the models when using word embedding against TF-IDF regarding all metrics analyzed. This improvement is especially significant for RNN, which is the method providing the best results when used in conjunction with word embedding. On the contrary, RNN is the worst of the three analyzed methods when used with TF-IDF. Figure 14 shows the metric values yielded by the RNN models.

We can also see in the graphs above that there are no significant differences in the values of the evaluation measures between the three deep learning techniques in the case of word embedding, while in the case of TF-IDF the differences in the results of the three methods are important.

Regarding the dataset size, its influence on the results is minimal for word embedding, being slightly greater and uneven for the TF-IDF method.

Therefore, from the analysis of the results obtained with the Sentiment140 dataset, we can deduce that word embedding is a more robust technique than TF-IDF. In addition, its use would allow us to work with a subset of data containing 50% or 60% of the total sample at a lower computational cost and with hardly any difference in the results.

After this preliminary study, we conducted additional experiments, in which the methods were tested on other datasets. Some of them contain tweets, while others contain different types of reviews, as noted in Section 4.1. These datasets contain significantly fewer examples than Sentiment140's, which has 1.6 million entries, so there was no problem working with the complete datasets.

Tables 2–6 show the results of the datasets; Figures 16–20 illustrate these results. These tables include the results of the complete Sentiment140 dataset, although as we found in the preliminary study, we could have used a subset of it and obtained similar results.

**Table 2.** Accuracy comparison for datasets with two classes (positive and negative).

| Datasets | TF-IDF | | | Word Embedding | | |
|---|---|---|---|---|---|---|
| | DNN | CNN | RNN | DNN | CNN | RNN |
| Sentiment140 | 0.76497407 | 0.76688544 | 0.56957939 | 0.78816761 | 0.80060849 | 0.82819948 |
| Tweets Airline | 0.85936944 | 0.85451457 | 0.82809226 | 0.8979309 | 0.90373439 | 0.90451624 |
| Tweets SemEval | 0.83674669 | 0.81377485 | 0.54857318 | 0.83674748 | 0.84313431 | 0.85172402 |
| IMDB Movie Reviews (1) | 0.85232000 | 0.82300000 | 0.56392000 | 0.84572000 | 0.86072000 | 0.87052000 |
| IMDB Movie Reviews (2) | 0.85512000 | 0.80628002 | 0.58724000 | 0.80252000 | 0.82624000 | 0.86688000 |
| Cornell Movie Reviews | 0.70437264 | 0.67867751 | 0.50787764 | 0.70221434 | 0.71365671 | 0.76693790 |
| Book Reviews | 0.75876443 | 0.72741509 | 0.5169437 | 0.74560455 | 0.76630924 | 0.73347052 |
| Music Reviews | 0.76850000 | 0.69200000 | 0.5170000 | 0.70800000 | 0.74450000 | 0.73100000 |

Table 3. The recall comparison for different datasets.

| Datasets | TF-IDF | | | Word Embedding | | |
|---|---|---|---|---|---|---|
| | DNN | CNN | RNN | DNN | CNN | RNN |
| Sentiment140 | 0.75775700 | 0.74076035 | 0.77731305 | 0.79096262 | 0.80080020 | 0.83692316 |
| Tweets Airline | 0.95565582 | 0.97003680 | 0.97417837 | 0.9577253 | **0.95924821** | **0.95086398** |
| Tweets SemEval | 0.80817204 | 0.7744086 | 0.09462366 | 0.80860215 | 0.81827957 | 0.83139785 |
| IMDB Movie Reviews (1) | 0.84072000 | 0.80080000 | 0.46880000 | 0.84360000 | 0.84960000 | 0.86808000 |
| IMDB Movie Reviews (2) | 0.87112000 | 0.75744000 | 0.56088000 | 0.78304000 | 0.83248000 | 0.88832000 |
| Cornell Movie Reviews | 0.71468474 | 0.67811554 | 0.84203575 | 0.70455552 | 0.72050860 | 0.80943813 |
| Book Reviews | 0.74221810 | 0.73009689 | 0.63040610 | 0.73912595 | 0.81599670 | 0.74824778 |
| Music Reviews | 0.76500000 | 0.69700000 | 0.74200000 | 0.68600000 | 0.72900000 | 0.73600000 |

Table 4. The precision comparison for different datasets.

| Datasets | TF-IDF | | | Word Embedding | | |
|---|---|---|---|---|---|---|
| | DNN | CNN | RNN | DNN | CNN | RNN |
| Sentiment140 | 0.75775700 | 0.74076035 | 0.77731305 | 0.79096262 | 0.80080020 | 0.83692316 |
| Tweets Airline | 0.88451273 | 0.86396543 | 0.83664149 | 0.91759076 | **0.92284682** | **0.93061436** |
| Tweets SemEval | 0.83504669 | 0.81594219 | 0.58839133 | 0.83492767 | 0.84024502 | 0.84745555 |
| IMDB Movie Reviews (1) | 0.85057402 | 0.83996428 | 0.61862397 | 0.84727512 | 0.8689903 | 0.87328478 |
| IMDB Movie Reviews (2) | 0.84410853 | 0.83943612 | 0.59209526 | 0.81478398 | 0.82221871 | 0.85179503 |
| Cornell Movie Reviews | 0.70070694 | 0.67920909 | 0.45431496 | 0.70142346 | 0.71117779 | 0.74808808 |
| Book Reviews | 0.77071809 | 0.72645030 | 0.56145983 | 0.74877856 | 0.74335207 | 0.73283058 |
| Music Reviews | 0.77097163 | 0.69126657 | 0.46068591 | 0.71900797 | 0.75328872 | 0.73186536 |

Table 5. The F-score comparison for different datasets.

| Datasets | TF-IDF | | | Word Embedding | | |
|---|---|---|---|---|---|---|
| | DNN | CNN | RNN | DNN | CNN | RNN |
| Sentiment140 | 0.76383225 | 0.75932297 | 0.64044056 | 0.78876610 | 0.80063705 | 0.82967613 |
| Tweets Airline | 0.91863362 | 0.91385701 | 0.90011208 | 0.93720980 | **0.94064543** | **0.94059646** |
| Tweets SemEval | 0.82114704 | 0.79433397 | 0.13751971 | 0.82130776 | 0.82884635 | 0.83874720 |
| IMDB Movie Reviews (1) | 0.85057402 | 0.81871110 | 0.46834558 | 0.84540045 | 0.85908973 | 0.87020187 |
| IMDB Movie Reviews (2) | 0.85740157 | 0.79633290 | 0.57606508 | 0.79859666 | 0.82731754 | 0.86967419 |
| Cornell Movie Reviews | 0.70731859 | 0.67852670 | 0.59007189 | 0.70290291 | 0.71560412 | 0.77594109 |
| Book Reviews | 0.75501388 | 0.72758940 | 0.51163296 | 0.74364502 | 0.77728796 | 0.73395298 |
| Music Reviews | 0.76770393 | 0.69126657 | 0.56736672 | 0.70080624 | 0.74026385 | 0.73207829 |

Table 6. The AUC comparison for different datasets.

| Datasets | TF-IDF | | | Word Embedding | | |
|---|---|---|---|---|---|---|
| | DNN | CNN | RNN | DNN | CNN | RNN |
| Sentiment140 | 0.76499683 | 0.76535951 | 0.56950939 | 0.78816189 | 0.80062146 | 0.82818031 |
| Tweets Airline | 0.73510103 | 0.68790047 | 0.61740993 | 0.81170789 | **0.82367939** | **0.83767632** |
| Tweets SemEval | 0.83484059 | 0.81115021 | 0.51834041 | 0.83487221 | 0.84147827 | 0.85037175 |
| IMDB Movie Reviews (1) | 0.85232000 | 0.82300000 | 0.56392000 | 0.84572000 | 0.86072000 | 0.87052000 |
| IMDB Movie Reviews (2) | 0.85512000 | 0.80628000 | 0.58724000 | 0.80252000 | 0.82624000 | 0.86688000 |
| Cornell Movie Reviews | 0.70437264 | 0.67867751 | 0.50787764 | 0.70221434 | 0.71365671 | 0.76693790 |
| Book Reviews | 0.75875593 | 0.72740157 | 0.51676458 | 0.74558854 | 0.76630592 | 0.73348794 |
| Music Reviews | 0.76850000 | 0.69200000 | 0.51700000 | 0.70800000 | 0.74450000 | 0.73207829 |

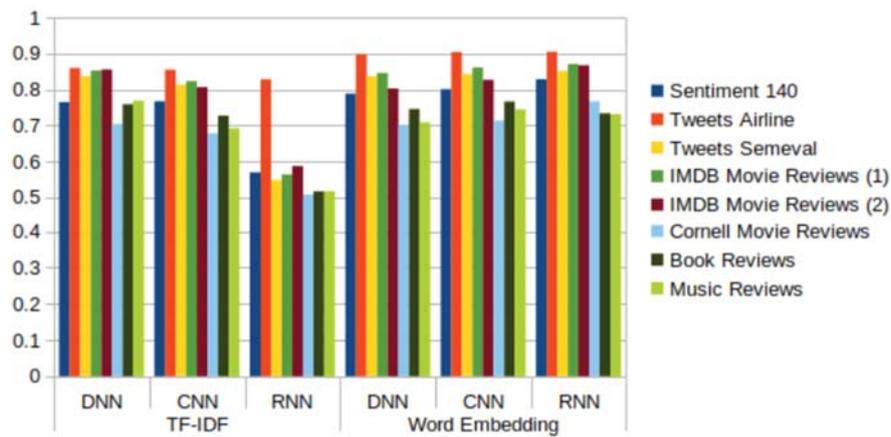

**Figure 16.** Accuracy values of deep-learning models with TF-IDF and word embedding for different datasets.

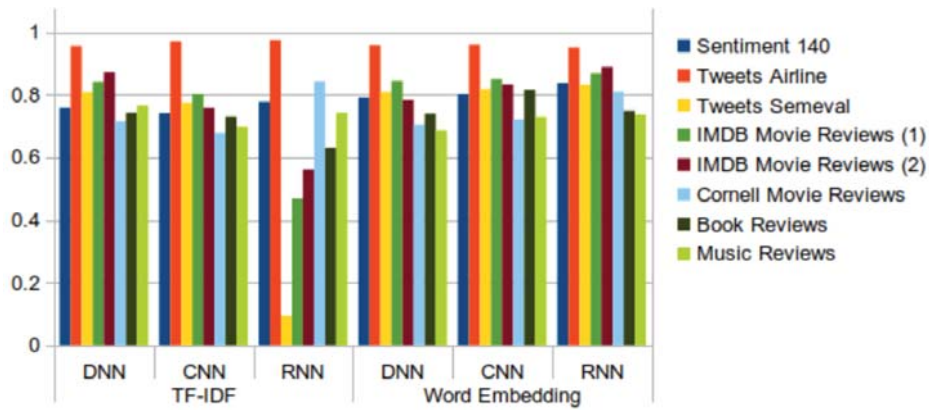

**Figure 17.** Recall values of DNN, CNN, and RNN models with TF-IDF and word embedding for different datasets.

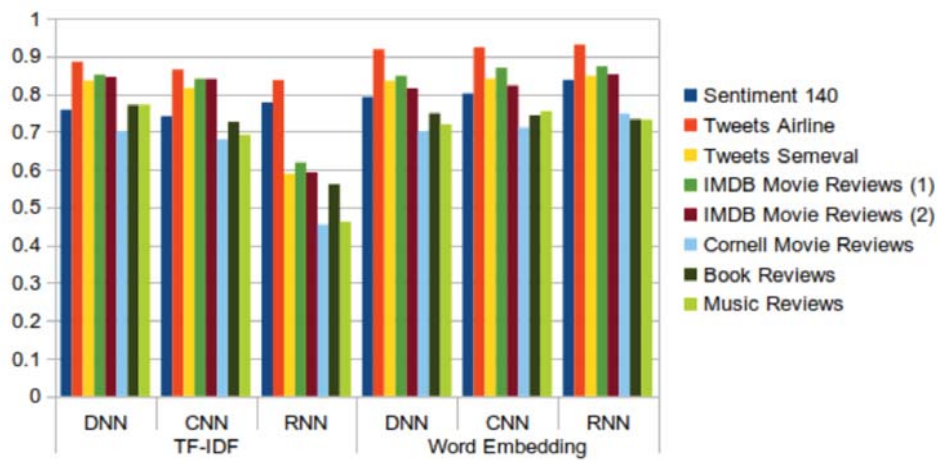

**Figure 18.** Precision values of DNN, CNN, and RNN models with TF-IDF and word embedding for different datasets.

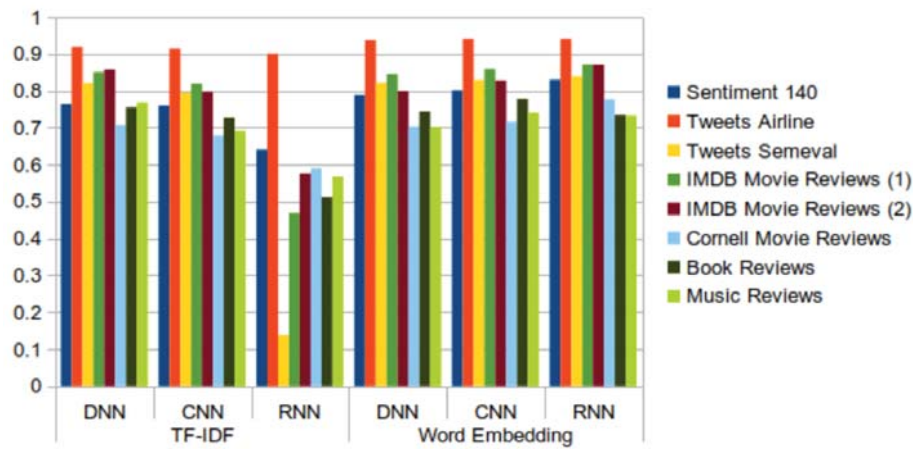

**Figure 19.** F-score values of DNN, CNN, and RNN models with TF-IDF and word embedding for different datasets.

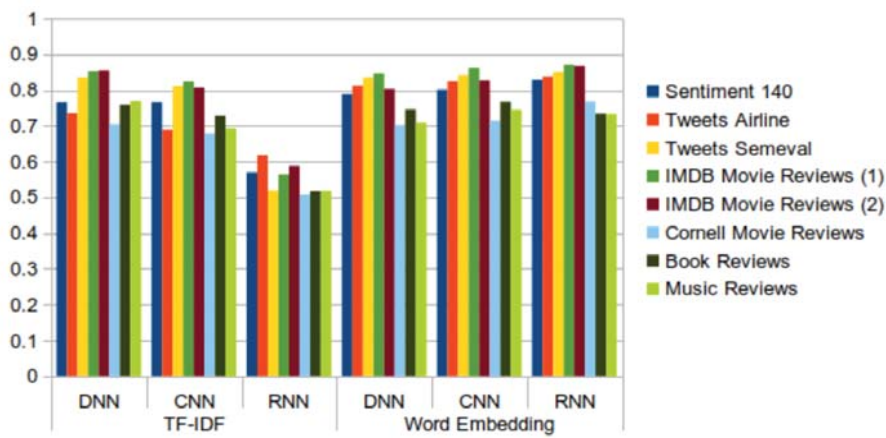

**Figure 20.** AUC values of DNN, CNN, and RNN models with TF-IDF and word embedding for different datasets.

The results obtained with the new datasets do nothing more than confirm the conclusions obtained after the analysis of the results of the Sentiment140 dataset. In general, the best behavior is shown by the combination of RNN and word embedding, although there are some exceptions. These are produced in the "book reviews" and "music review" datasets, where the values of all the metrics are slightly higher for DNN + TF-IDF than for RNN + word embedding. For "book reviews", the highest values for accuracy, recall, F-score, and AUC were given by CNN + word embedding. In addition, the Tweets Airline dataset is one of the datasets that shows the highest values for all metrics in all cases. We can also highlight that the recall metric shows an uneven behavior, especially for the model that combines RNN and TF-IDF. The same behavior was seen in the preliminary study with the Sentiment140 dataset (Figure 11). Likewise, as in the preliminary study, we can affirm that word embedding is a more appropriate technique than TF-IDF for performing sentiment analysis, despite the slight improvements obtained with TF-IDF for some data sets.

After analyzing the results concerning the quality of the predictions, it is necessary to obtain information on the computational cost associated with the induction of the models, since the differences between the results or some of them are not very significant. The aim is to know the extent to which the best reliability values are obtained at the expense of a higher or lower computational cost.

The CPU times are shown in Tables 7 and 8. Table 7 shows the processing time required to induce the models from the Sentiment140 dataset and its subsets. Table 8 contains the CPU time required for all datasets involved in the experiments.

Table 7. CPU times of experiments with numbers of tweets with a GPU.

| Dataset (%) | TF-IDF | | | Word Embedding | | |
|---|---|---|---|---|---|---|
| | DNN | CNN | RNN | DNN | CNN | RNN |
| 10 | 1 min 37 s | 1 min 14 s | 11 min 18 s | 25.8 s | 39.6 s | 4 min 58 s |
| 20 | 2 min 32 s | 2 min 25 s | 22 min 14 s | 41.3 s | 1 min 18 s | 11 min 59 s |
| 30 | 3 min 26 s | 3 min 34 s | 32 min 56 s | 1 min | 1 min 53 s | 18 min 57 s |
| 40 | 4 min 19 s | 4 min 53 s | 44 min 1 s | 1 min 21 s | 2 min 32 s | 25 min 9 s |
| 50 | **5 min 12 s** | **6 min 9 s** | **54 min 32 s** | **1 min 44 s** | **3 min 10 s** | **31 min 29 s** |
| 60 | 6 min 33 s | 7 min 23 s | 1h 5 min 26 s | 2 min 10 s | 3 min 52 s | 37 min 35 s |
| 70 | 7 min 47 s | 10 min 20 s | 1 h15 min5 s | 2 min 45 s | 4 min 38 s | 44 min 16 s |
| 80 | 9 min 4 s | 18 min 32 s | 1 h27 min22 s | 3 min 19 s | 5 min 31 s | 50 min 47 s |
| 90 | 10 min 14 s | 29 min 49 s | 1 h37 min59 s | 3 min 47 s | 6 min 12 s | 57 min 3 s |
| 100 | **11 min 55 s** | **38 min 17 s** | **1 h48 min52 s** | **4 min 18 s** | **7 min 3 s** | **1 h 4 min 16 s** |

Table 8. CPU times of experiments with numbers of datasets with a GPU.

| Dataset | TF-IDF | | | Word Embedding | | |
|---|---|---|---|---|---|---|
| | DNN | CNN | RNN | DNN | CNN | RNN |
| Sentiment140 | 11 min 55 s | 3 8min 17 s | 1h48 min 52 s | 4 min 18 s | 7 min 3 s | 1 h 4 min 16 s |
| Tweets Airline | 1 min | 34.41 s | 1 h 54 s | 30.66 s | 1 min 22 s | 2 min 41 s |
| Tweets SemEval | 20.53 s | 24.5 s | 23 min 52 s | 26.75 s | 1 min 11 s | 2 min 43 s |
| IMDB Movie Reviews (1) | 1 min 11 s | 1 min 7 s | 1 h25 min 48 s | 21.13 s | 32.66 s | 7 min 42 s |
| IMDB Movie Reviews (2) | 17.78 s | 22.05 s | 30 min 21 s | 31.32 s | 36.81 s | 8 min 23 s |
| Cornell Movie Reviews | 23.2 s | 16.83 s | 31 min 55 s | 12.9 s | 21.26 s | 4 min 40 s |
| Book Reviews | 11.93 s | 10.12 s | 21 min 9 s | 16.21 s | 20.6 s | 2 min17 s |
| Music Reviews | 26.48 s | 17.35 s | 29 min 50 s | 13.94 s | 16.89 s | 4 min 42 s |

The tables show that the use of TF-IDF, which produces less reliable models, requires longer computational time than the use of word embedding. This is one more reason to consider this last technique as the most recommendable. However, RNN is the most time-consuming algorithm, both with TF-IDF and with word embedding. Given that the improvements of RNN with respect to DNN and CNN are not very significant in the latter case, the use of these two methods could be considered more appropriate when the computational cost needs to be reduced.

Regarding the large Sentiment140 dataset shown in Table 8, if the sample size is reduced by 50%, the evaluation measures are not significantly affected, but the processing time is reduced by 50%.

Moving to a comparison between the DNN and CNN models, CNN has slightly longer processing times, but it also has much better evaluation measures than DNN.

From the analysis of overall accuracy, recall, precision, F-scores, AUC values, and CPU times, we highlight some patterns for high and low performance of the sentiment analysis methods. We are aware that different types of datasets influence the results of a sentiment analysis differently [76].

- The DNN model is simple to implement and provides results within a short period of time—around 1 min for the majority of datasets, except dataset Sentiment140, for which the model took 12 min to obtain the results. Although the model is quick to train, the overall accuracy of the model is average (around 75% to 80%) in all of the tested datasets, including tweets and reviews.
- The CNN model is also fast to train and test, although possibly a bit slower than DNN. The model offers higher accuracy (over 80%) on both tweet and review datasets.
- The RNN model has the highest reliability when word embedding is applied, however its computational time is also the highest. When using RNN with TF-IDF, it takes a longer time than other models and results in lower accuracy (around 50%) in the sentiment analysis of tweet and review datasets.

In comparative studies presented in [5,17,19,20], which were performed by using tweets or reviews datasets, the evaluation of results was made only in terms of accuracy, however the processing time was not considered. Regarding the continuously expanding size and complexity of big data in the future, it is crucial to consider both reliability and time, especially in critical systems requiring a fast response [77]. In this work, two techniques (TF-IDF and word embedding) are examined on three deep learning algorithms, which give an extended overview of performances of sentiment analysis using deep learning techniques.

Finally, general summaries of the results archived in the experiments referenced earlier are explained below:

- Three deep learning models (DNN, CNN, and RNN) were used to perform sentiment analysis experiments. The CNN model was found to offer the best tradeoff between the processing time and the accuracy of results. Although the RNN model had the highest degree of accuracy when used with word embedding, its processing time was 10 times longer than that of the CNN model. The RNN model is not effective when used with the TF-IDF technique, and its far higher processing time leads to results that are not significantly better. DNN is a simple deep learning model that has average processing times and yields average results. Future research on deep learning models can focus on ways of improving the tradeoff between the accuracy of results and the processing times.
- Related techniques (TF-IDF and word embedding) are used to transfer text data (tweets, reviews) into a numeric vector before feeding them into a deep learning model. The results when TF-IDF is used are poorer than when word embedding is used. Moreover, the TF-IDF technique used with the RNN model takes has a longer processing time and yields less reliable results. However, when RNN is used with word embedding, the results are much better. Future work can explore how to improve these and other techniques to achieve even better results.
- The results from the datasets containing tweets and IMDB movie review datasets are better than the results from the other datasets containing reviews. Regarding tweets data, the models induced from the Tweets Airline dataset, focused on a specific topic, show better performance than those built from datasets about generic topics.

**6. Conclusions**

In this paper, we described the core of deep learning models and related techniques that have been applied to sentiment analysis for social network data. We used word embedding and TF-IDF to transform input data before feeding that data into deep learning models. The architectures of DNN, CNN, and RNN were analyzed and combined with word embedding and TF-IDF to perform sentiment analysis. We conducted some experiments to evaluate DNN, CNN, and RNN models on datasets of different topics, including tweets and reviews. We also discussed related research in the field. This information, combined with the results of our experiments, gives us a broad perspective on applying deep learning models for sentiment analysis, as well as combining these models with text preprocessing techniques.

After the analysis of 32 papers, DNN, CNN, and hybrid approaches were identified as the most widely used models for sentiment polarity analysis. Another conclusion extracted from the analysis was the fact that common techniques, such as CNN, RNN, and LSTM, are individually tested in these studies on different datasets, however there is a lack of a comparative analysis for them. In addition, the results presented in most papers are given in terms of reliability, without considering computational time.

The experiments conducted in this work were designed to help fill the gaps mentioned above. We studied the impacts of different types of datasets, feature extraction techniques, and deep learning models, with a special focus on the problem of sentiment polarity analysis. The results show that it is better to combine deep learning techniques with word embedding than with TF-IDF when performing a sentiment analysis. The experiments also revealed that CNN outperforms other models, presenting a good balance between accuracy and CPU runtime. RNN reliability is slightly higher than CNN reliability with most datasets but its computational time is much longer. One last conclusion derived from the study is the observation that the effectiveness of the algorithms depends largely on the characteristics of the datasets, hence the convenience of testing deep learning methods with more datasets in order to cover a greater diversity of characteristics.

In future work, we will focus on exploring hybrid approaches, where multiple models and techniques are combined in order to enhance the sentiment classification accuracy achieved by the individual models or techniques, as well as to reduce the computational cost. The aim is to extend the comparative study to include both new methods and new types of data. Therefore, the reliability and processing time of hybrid models will be evaluated with several types of data, such as status, comments, and news on social media. We will also intend to address the problem of aspect sentiment analysis in order to gain deeper insight into user sentiments by associating them with specific features or topics. This has great relevance for many companies, since it allows them to obtain detailed feedback from users, and thus know which aspects of their products or services should be improved.

**Author Contributions:** Conceptualization, N.C.D. and M.N.M.-G.; methodology, M.N.M.-G.; software, N.C.D.; validation, M.N.M.-G., and F.D.l.P.; formal analysis, N.C.D., and M.N.M.-G.; investigation, N.C.D.; data curation, N.C.D.; writing—original draft preparation, N.C.D.; writing—review and editing, M.N.M.-G. and F.D.l.P.; visualization, N.C.D.; supervision, M.N.M.-G. and F.D.l.P.; project administration, F.D.l.P.; funding acquisition, F.D.l.P. All authors have read and agreed to the published version of the manuscript.

**Funding:** This work was supported by the Spanish government and European (Fondo Europeo de Desarrollo Regional) FEDER funds, project InEDGEMobility: Movilidad inteligente y sostenible soportada por Sistemas Multi-agentes y Edge Computing (RTI2018-095390-B-C32).

**Conflicts of Interest:** The authors declare no conflict of interest. The funders had no role in the design of the study; in the collection, analyses, or interpretation of data; in the writing of the manuscript, or in the decision to publish the results.

## References

1. Pouli, V.; Kafetzoglou, S.; Tsiropoulou, E.E.; Dimitriou, A.; Papavassiliou, S. Personalized multimedia content retrieval through relevance feedback techniques for enhanced user experience. In Proceedings of the 2015 13th International Conference on Telecommunications (ConTEL), Graz, Austria, 13–15 July 2015; pp. 1–8.
2. Thai, M.T.; Wu, W.; Xiong, H. *Big Data in Complex and Social Networks*; CRC Press: Boca Raton, FL, USA, 2016.
3. Cambria, E.; Das, D.; Bandyopadhyay, S.; Feraco, A. *A Practical Guide to Sentiment Analysis*; Springer: Berlin, Germany, 2017.
4. Hussein, D.M.E.-D.M. A survey on sentiment analysis challenges. *J. King Saud Univ. Eng. Sci.* **2018**, *30*, 330–338. [CrossRef]
5. Sohangir, S.; Wang, D.; Pomeranets, A.; Khoshgoftaar, T.M. Big Data: Deep Learning for financial sentiment analysis. *J. Big Data* **2018**, *5*, 3. [CrossRef]
6. Jangid, H.; Singhal, S.; Shah, R.R.; Zimmermann, R. Aspect-Based Financial Sentiment Analysis using Deep Learning. In Proceedings of the Companion of the The Web Conference 2018 on The Web Conference, Lyon, France, 23–27 April 2018; pp. 1961–1966.
7. Keenan, M.J.S. *Advanced Positioning, Flow, and Sentiment Analysis in Commodity Markets*; Wiley: Hoboken, NJ, USA, 2018.
8. Satapathy, R.; Cambria, E.; Hussain, A. *Sentiment Analysis in the Bio-Medical Domain*; Springer: Berlin, Germany, 2017.
9. Rajput, A. Natural Language Processing, Sentiment Analysis, and Clinical Analytics. In *Innovation in Health Informatics*; Elsevier: Amsterdam, The Netherlands, 2020; pp. 79–97.
10. Qian, J.; Niu, Z.; Shi, C. Sentiment Analysis Model on Weather Related Tweets with Deep Neural Network. In Proceedings of the 2018 10th International Conference on Machine Learning and Computing, Macau, China, 26–28 February 2018; pp. 31–35.
11. Pham, D.-H.; Le, A.-C. Learning multiple layers of knowledge representation for aspect based sentiment analysis. *Data Knowl. Eng.* **2018**, *114*, 26–39. [CrossRef]
12. Preethi, G.; Krishna, P.V.; Obaidat, M.S.; Saritha, V.; Yenduri, S. Application of deep learning to sentiment analysis for recommender system on cloud. In Proceedings of the 2017 International Conference on Computer, Information and Telecommunication Systems (CITS), Dalian, China, 21–23 July 2017; pp. 93–97.

13. Ain, Q.T.; Ali, M.; Riaz, A.; Noureen, A.; Kamran, M.; Hayat, B.; Rehman, A. Sentiment analysis using deep learning techniques: A review. *Int. J. Adv. Comput. Sci. Appl.* **2017**, *8*, 424.
14. Gao, Y.; Rong, W.; Shen, Y.; Xiong, Z. Convolutional neural network based sentiment analysis using Adaboost combination. In Proceedings of the 2016 International Joint Conference on Neural Networks (IJCNN), Vancouver, BC, Canada, 24–29 July 2016; pp. 1333–1338.
15. Hassan, A.; Mahmood, A. Deep learning approach for sentiment analysis of short texts. In Proceedings of the Third International Conference on Control, Automation and Robotics (ICCAR), Nagoya, Japan, 24–26 April 2017; pp. 705–710.
16. Kraus, M.; Feuerriegel, S. Sentiment analysis based on rhetorical structure theory: Learning deep neural networks from discourse trees. *Expert Syst. Appl.* **2019**, *118*, 65–79. [CrossRef]
17. Li, L.; Goh, T.-T.; Jin, D. How textual quality of online reviews affect classification performance: A case of deep learning sentiment analysis. *Neural Comput. Appl.* **2018**, 1–29. [CrossRef]
18. Singhal, P.; Bhattacharyya, P. *Sentiment Analysis and Deep Learning: A Survey*; Center for Indian Language Technology, Indian Institute of Technology: Bombay, Indian, 2016.
19. Alharbi, A.S.M.; de Doncker, E. Twitter sentiment analysis with a deep neural network: An enhanced approach using user behavioral information. *Cogn. Syst. Res.* **2019**, *54*, 50–61. [CrossRef]
20. Abid, F.; Alam, M.; Yasir, M.; Li, C.J. Sentiment analysis through recurrent variants latterly on convolutional neural network of Twitter. *Future Gener. Comput. Syst.* **2019**, *95*, 292–308. [CrossRef]
21. Aggarwal, C.C. *Neural Networks and Deep Learning*; Springer: Berlin, Germany, 2018.
22. Zhang, L.; Wang, S.; Liu, B. Deep learning for sentiment analysis: A survey. *WIREs Data Min. Knowl. Discov.* **2018**, *8*, e1253. [CrossRef]
23. Britz, D. Recurrent Neural Networks Tutorial, Part 1–Introduction to Rnns. Available online: http://www.wildml.com/2015/09/recurrent-neural-networkstutorial-part-1-introduction-to-rnns/ (accessed on 12 March 2020).
24. Hochreiter, S.; Schmidhuber, J. LSTM can solve hard long time lag problems. In Proceedings of the Advances in Neural Information Processing Systems, Denver, CO, USA, 2–5 December 1996; pp. 473–479.
25. Ruangkanokmas, P.; Achalakul, T.; Akkarajitsakul, K. Deep Belief Networks with Feature Selection for Sentiment Classification. In Proceedings of the 2016 7th International Conference on Intelligent Systems, Modelling and Simulation (ISMS), Bangkok, Thailand, 25–27 January 2016; pp. 9–14.
26. Socher, R.; Lin, C.C.; Manning, C.; Ng, A.Y. Parsing natural scenes and natural language with recursive neural networks. In Proceedings of the 28th International Conference on Machine Learning (ICML-11), Bellevue, WA, USA, 28 June–2 July 2011; pp. 129–136.
27. Long, H.; Liao, B.; Xu, X.; Yang, J. A hybrid deep learning model for predicting protein hydroxylation sites. *Int. J. Mol. Sci.* **2018**, *19*, 2817. [CrossRef]
28. Vateekul, P.; Koomsubha, T. A study of sentiment analysis using deep learning techniques on Thai Twitter data. In Proceedings of the 2016 13th International Joint Conference on Computer Science and Software Engineering (JCSSE), Khon Kaen, Thailand, 13–15 July 2016; pp. 1–6.
29. Ghosh, R.; Ravi, K.; Ravi, V. A novel deep learning architecture for sentiment classification. In Proceedings of the 2016 3rd International Conference on Recent Advances in Information Technology (RAIT), Dhanbad, India, 3–5 March 2016; pp. 511–516.
30. Bhavitha, B.; Rodrigues, A.P.; Chiplunkar, N.N. Comparative study of machine learning techniques in sentimental analysis. In Proceedings of the 2017 International Conference on Inventive Communication and Computational Technologies (ICICCT), Coimbatore, India, 10–11 March 2017; pp. 216–221.
31. Salas-Zárate, M.P.; Medina-Moreira, J.; Lagos-Ortiz, K.; Luna-Aveiga, H.; Rodriguez-Garcia, M.A.; Valencia-García, R.J.C. Sentiment analysis on tweets about diabetes: An aspect-level approach. *Comput. Math. Methods Med.* **2017**, *2017*. [CrossRef] [PubMed]
32. Huq, M.R.; Ali, A.; Rahman, A. Sentiment analysis on Twitter data using KNN and SVM. *IJACSA Int. J. Adv. Comput. Sci. Appl.* **2017**, *8*, 19–25.
33. Pinto, D.; McCallum, A.; Wei, X.; Croft, W.B. Table extraction using conditional random fields. In Proceedings of the 26th Annual International ACM SIGIR Conference on Research and Development in Informaion Retrieval, Toronto, ON, Canada, 28 July–1 August 2003; pp. 235–242.

34. Soni, S.; Sharaff, A. Sentiment analysis of customer reviews based on hidden markov model. In Proceedings of the 2015 International Conference on Advanced Research in Computer Science Engineering & Technology (ICARCSET 2015), Unnao, India, 6 March 2015; pp. 1–5.
35. Zhang, X.; Zheng, X. Comparison of Text Sentiment Analysis Based on Machine Learning. In Proceedings of the 2016 15th International Symposium on Parallel and Distributed Computing (ISPDC), Fuzhou, China, 8–10 July 2016; pp. 230–233.
36. Malik, V.; Kumar, A. Communication. Sentiment Analysis of Twitter Data Using Naive Bayes Algorithm. *Int. J. Recent Innov. Trends Comput. Commun.* **2018**, *6*, 120–125.
37. Mehra, N.; Khandelwal, S.; Patel, P. *Sentiment Identification Using Maximum Entropy Analysis of Movie Reviews*; Stanford University: Stanford, CA, USA, 2002.
38. Wu, H.; Li, J.; Xie, J. Maximum entropy-based sentiment analysis of online product reviews in Chinese. In *Automotive, Mechanical and Electrical Engineering*; CRC Press: Boca Raton, FL, USA, 2017; pp. 559–562.
39. Firmino Alves, A.L.; Baptista, C.d.S.; Firmino, A.A.; Oliveira, M.G.d.; Paiva, A.C.D. A Comparison of SVM versus naive-bayes techniques for sentiment analysis in tweets: A case study with the 2013 FIFA confederations cup. In Proceedings of the 20th Brazilian Symposium on Multimedia and the Web, João Pessoa, Brazil, 18–21 November 2014; pp. 123–130.
40. Pandey, A.C.; Rajpoot, D.S.; Saraswat, M. Twitter sentiment analysis using hybrid cuckoo search method. *Inf. Process. Manag.* **2017**, *53*, 764–779. [CrossRef]
41. Medhat, W.; Hassan, A.; Korashy, H. Sentiment analysis algorithms and applications: A survey. *Ain Shams Eng. J.* **2014**, *5*, 1093–1113. [CrossRef]
42. Mikolov, T.; Sutskever, I.; Chen, K.; Corrado, G.S.; Dean, J. Distributed representations of words and phrases and their compositionality. In Proceedings of the Advances in neural Information Processing Systems, Lake Tahoe, NV, USA, 5–10 December 2013; pp. 3111–3119.
43. Jain, A.P.; Dandannavar, P. Application of machine learning techniques to sentiment analysis. In Proceedings of the 2016 2nd International Conference on Applied and Theoretical Computing and Communication Technology (iCATccT), Karnataka, India, 21–23 July 2016; pp. 628–632.
44. Tang, D.; Qin, B.; Liu, T. Deep learning for sentiment analysis: Successful approaches and future challenges. *Wiley Interdiscip. Rev. Data Min. Knowl. Discov.* **2015**, *5*, 292–303. [CrossRef]
45. Sharef, N.M.; Zin, H.M.; Nadali, S. Overview and Future Opportunities of Sentiment Analysis Approaches for Big Data. *JCS* **2016**, *12*, 153–168. [CrossRef]
46. Rojas-Barahona, L.M. Deep learning for sentiment analysis. *Lang. Linguist. Compass* **2016**, *10*, 701–719. [CrossRef]
47. Roshanfekr, B.; Khadivi, S.; Rahmati, M. Sentiment analysis using deep learning on Persian texts. In Proceedings of the 2017 Iranian Conference on Electrical Engineering (ICEE), Tehran, Iran, 2–4 May 2017; pp. 1503–1508.
48. Jeong, B.; Yoon, J.; Lee, J.-M. Social media mining for product planning: A product opportunity mining approach based on topic modeling and sentiment analysis. *Int. J. Inf. Manag.* **2019**, *48*, 280–290. [CrossRef]
49. Gupta, U.; Chatterjee, A.; Srikanth, R.; Agrawal, P. A sentiment-and-semantics-based approach for emotion detection in textual conversations. *arXiv* **2017**, arXiv:1707.06996.
50. Ramadhani, A.M.; Goo, H.S. Twitter sentiment analysis using deep learning methods. In Proceedings of the 2017 7th International Annual Engineering Seminar (InAES), Yogyakarta, Indonesia, 1–2 August 2017; pp. 1–4.
51. Paredes-Valverde, M.A.; Colomo-Palacios, R.; Salas-Zárate, M.D.P.; Valencia-García, R. Sentiment analysis in Spanish for improvement of products and services: A deep learning approach. *Sci. Program.* **2017**, *2017*. [CrossRef]
52. Yang, C.; Zhang, H.; Jiang, B.; Li, K.J. Aspect-based sentiment analysis with alternating coattention networks. *Inf. Process. Manag.* **2019**, *56*, 463–478. [CrossRef]
53. Do, H.H.; Prasad, P.; Maag, A.; Alsadoon, A.J. Deep Learning for Aspect-Based Sentiment Analysis: A Comparative Review. *Expert Syst. Appl.* **2019**, *118*, 272–299. [CrossRef]
54. Schmitt, M.; Steinheber, S.; Schreiber, K.; Roth, B. Joint Aspect and Polarity Classification for Aspect-based Sentiment Analysis with End-to-End Neural Networks. *arXiv* **2018**, arXiv:1808.09238.
55. Balabanovic, M.; Shoham, Y. Combining content-based and collaborative recommendation. *Commun. ACM* **1997**, *40*, 66–72. [CrossRef]

56. Wang, Y.; Wang, M.; Xu, W. A sentiment-enhanced hybrid recommender system for movie recommendation: A big data analytics framework. *Wirel. Commun. Mob. Comput.* **2018**, *2018*. [CrossRef]
57. Singh, V.K.; Mukherjee, M.; Mehta, G.K. Combining collaborative filtering and sentiment classification for improved movie recommendations. In Proceedings of the International Workshop on Multi-disciplinary Trends in Artificial Intelligence, Hyderabad, India, 7–9 December 2011; pp. 38–50.
58. Chen, Z.; Liu, B. Lifelong machine learning. *Synth. Lect. Artif. Intell. Mach. Learn.* **2018**, *12*, 1–207. [CrossRef]
59. Stai, E.; Kafetzoglou, S.; Tsiropoulou, E.E.; Papavassiliou, S.J. A holistic approach for personalization, relevance feedback & recommendation in enriched multimedia content. *Multimed. Tools Appl.* **2018**, *77*, 283–326.
60. Wu, C.; Wu, F.; Wu, S.; Yuan, Z.; Liu, J.; Huang, Y. Semi-supervised dimensional sentiment analysis with variational autoencoder. *Knowl. Based Syst.* **2019**, *165*, 30–39. [CrossRef]
61. Zhang, Z.; Zou, Y.; Gan, C. Textual sentiment analysis via three different attention convolutional neural networks and cross-modality consistent regression. *Neurocomputing* **2018**, *275*, 1407–1415. [CrossRef]
62. Tang, D.; Zhang, M. Deep Learning in Sentiment Analysis. In *Deep Learning in Natural Language Processing*; Springer: Berlin, Germany, 2018; pp. 219–253.
63. Araque, O.; Corcuera-Platas, I.; Sanchez-Rada, J.F.; Iglesias, C.A. Enhancing deep learning sentiment analysis with ensemble techniques in social applications. *Expert Syst. Appl.* **2017**, *77*, 236–246. [CrossRef]
64. Liu, J.; Chang, W.-C.; Wu, Y.; Yang, Y. Deep learning for extreme multi-label text classification. In Proceedings of the 40th International ACM SIGIR Conference on Research and Development in Information Retrieval, Tokyo, Japan, 7–11 August 2017; pp. 115–124.
65. Chen, M.; Wang, S.; Liang, P.P.; Baltrušaitis, T.; Zadeh, A.; Morency, L.-P. Multimodal sentiment analysis with word-level fusion and reinforcement learning. In Proceedings of the 19th ACM International Conference on Multimodal Interaction, Glasgow, UK, 13–17 November 2017; pp. 163–171.
66. Al-Sallab, A.; Baly, R.; Hajj, H.; Shaban, K.B.; El-Hajj, W.; Badaro, G. Aroma: A recursive deep learning model for opinion mining in arabic as a low resource language. *ACM Trans. Asian Low-Resour. Lang. Inf. Process. TALLIP* **2017**, *16*, 1–20. [CrossRef]
67. Kumar, S.; Gahalawat, M.; Roy, P.P.; Dogra, D.P.; Kim, B.-G.J.E. Exploring Impact of Age and Gender on Sentiment Analysis Using Machine Learning. *Electronics* **2020**, *9*, 374. [CrossRef]
68. Available online: http://help.sentiment140.com/site-functionality (accessed on 12 March 2020).
69. Available online: https://www.kaggle.com/crowdflower/twitter-airline-sentiment (accessed on 12 March 2020).
70. Available online: http://alt.qcri.org/semeval2017/ (accessed on 12 March 2020).
71. Available online: https://www.kaggle.com/c/word2vec-nlp-tutorial/data (accessed on 12 March 2020).
72. Maas, A.L.; Daly, R.E.; Pham, P.T.; Huang, D.; Ng, A.Y.; Potts, C. Learning word vectors for sentiment analysis. In Proceedings of the 49th Annual Meeting of the Association for Computational Linguistics: Human Language Technologies-Volume 1, Portland, OR, USA, 19–24 June 2011; pp. 142–150.
73. Available online: http://www.cs.cornell.edu/people/pabo/movie-review-data/ (accessed on 12 March 2020).
74. Blitzer, J.; Dredze, M.; Pereira, F. Biographies, bollywood, boom-boxes and blenders: Domain adaptation for sentiment classification. In Proceedings of the 45th Annual Meeting of the Association of Computational Linguistics, Prague, Czech Republic, 23–30 June 2007; pp. 440–447.
75. Kim, Y.; Sidney, J.; Buus, S.; Sette, A.; Nielsen, M.; Peters, B. Dataset size and composition impact the reliability of performance benchmarks for peptide-MHC binding predictions. *BMC Bioinform.* **2014**, *15*, 241. [CrossRef]
76. Choi, Y.; Lee, H.J. Data properties and the performance of sentiment classification for electronic commerce applications. *Inf. Syst. Front.* **2017**, *19*, 993–1012. [CrossRef]
77. Neppalli, V.K.; Caragea, C.; Squicciarini, A.; Tapia, A.; Stehle, S.J. Sentiment analysis during Hurricane Sandy in emergency response. *Int. J. Disaster Risk Reduct.* **2017**, *21*, 213–222. [CrossRef]